\documentclass[10pt,twocolumn,letterpaper]{article}
\usepackage{cvpr}
\usepackage{times}
\usepackage{epsfig}
\usepackage{wrapfig, blindtext}
\usepackage{graphicx}
\usepackage{amsmath}
\usepackage{cite}
\usepackage{amssymb}
\usepackage{fancyhdr}
\usepackage{parskip}
\usepackage{lettrine}
\usepackage{tikz}
\usetikzlibrary{arrows}
\usetikzlibrary{arrows.meta}
\usepackage{subcaption}
\usepackage{color}
\usepackage{xifthen}

\newcommand{\comment}[1]{}

\newcommand{\expectation}[2]{\mathbb{E}_{ #2 }\left[#1 \right]}
\newcommand{\parentheses}[1]{\left( #1 \right)}

\newcommand{\tensor}[4]
{
    \ifthenelse {\equal{#2}{}} 
        {\mathbf{#1}^{#3}_{#4}} 
        {\mathbf{#1}\scalebox{.5}{\{$#2$\}}^{#3}_{#4}} 
}
\usepackage[breaklinks=true,bookmarks=false]{hyperref}
\cvprfinalcopy % *** Uncomment this line for the final submission
\def\cvprPaperID{****} % *** Enter the CVPR Paper ID here
\begin{document}
\title{Context-sensitive neocortical neurons transform the effectiveness and efficiency of neural information processing}

% \title{Short Title: Context-sensitive neural information processing}

\author{Ahsan Adeel$^{1,2,3}$\thanks{$^1$Oxford Computational Neuroscience Lab, Nuffield Department of Surgical Sciences, University of Oxford, Oxford, UK.
$^2$CMI Lab, University of Wolverhampton, Wolverhampton, UK. $^3$deepCI.org, Parkside Terrace, Edinburgh, UK. Email: ahsan.adeel@deepci.org}, Mario Franco$^2$, Mohsin Raza$^2$,Khubaib Ahmed$^2$ }

\maketitle

\begin{abstract} 
Deep learning (DL) has big-data processing capabilities that are as good, or even better, than those of humans in many real-world domains, but at the cost of high energy requirements that may be unsustainable in some applications and of errors, that, though infrequent, can be large. We hypothesise that a fundamental weakness of DL lies in its intrinsic dependence on integrate-and-fire point neurons that maximise information transmission irrespective of whether it is relevant in the current context or not. This leads to unnecessary neural firing and to the feedforward transmission of conflicting messages, which makes learning difficult and processing energy inefficient. Here we show how to circumvent these limitations by mimicking the capabilities of context-sensitive neocortical neurons that receive input from diverse sources as a context to amplify and attenuate the transmission of relevant and irrelevant information, respectively. Our results show that, in the case of audio-visual processing, nets composed of context-sensitive local processors can use video information as a context that guides audio signal processing towards the currently relevant information far more effectively and efficiently than current forms of DL.
\end{abstract}
% \\\textbf{Teaser:} \textit{\textbf{Context-sensitive neural information processing inferred from cellular discoveries does indeed have the big data processing abilities hypothesised for them.}} \\

\begin{figure} 
	\centering
	\includegraphics[trim=0cm 0cm 0cm 0cm, clip=true, width=0.35\textwidth]{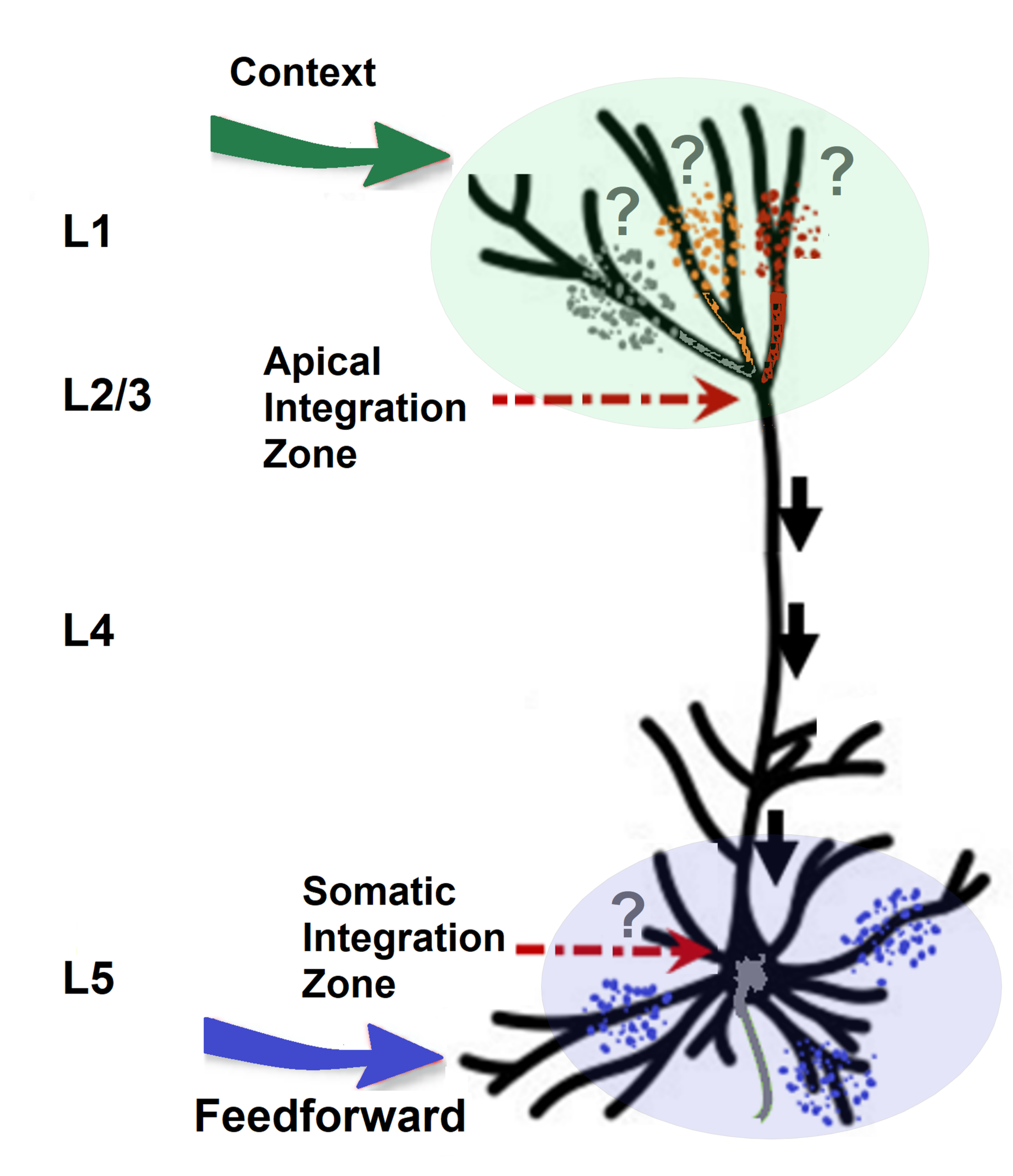}
	\caption{Context-sensitive neocortical neuron whose apical dendrites are in layer 1 (L1) with cell body and basal dendrites in deeper layers. The apical tuft receives input from diverse sources as context to amplify the transmission of coherent feedforward signals. However, to make this mechanism process large-scale
complex real-world data effectively and efficiently, it is crucial to understand different kinds of information that arrive at the apical tuft and how they influence the cell’s response to the feedforward input.}
	\label{l5pc}
 \vspace{-1em}
\end{figure}
\textbf{Introduction
}
\\For more than a century, theories of brain function have seen pyramidal cells as integrate-and-fire `point' neurons that integrate all the incoming synaptic inputs in an identical way to compute a net level of cellular activation 
\cite{hausser2001synaptic,burkitt2006review}. Modern DL models \cite{lecun2015deep} and their hardware implementations (e.g. \cite{davies2018loihi, merolla2014million, furber2014spinnaker, benjamin2014neurogrid, lichtsteiner2008128, moradi2017scalable, thakur2018large, qiao2015reconfigurable, valentian2019fully, wang2017neuromorphic, pei2019towards, frenkel20180, chen20184096}), inspired by the point neuron, have demonstrated ground-breaking performance improvements in a range of real-world problems, including speech processing, image recognition, and object detection, yet their energy demand and complexity scale so rapidly that the technology often becomes economically, technically, and environmentally unsustainable \cite{mehonic2022brain, thompson2020computational, strubell2019energy, strubell2020energy}. Attempts to solve energy issues in DL models have shown efficient computing \cite{chen2018sparse, hoefler2021sparsity, makhzani2015winner, kurtz2020inducing, mocanu2018scalable, ahmad2019can, changpinyo2017power, gale2020sparse}, though a biologically plausible solution which can achieve human-level computational efficiency remains an open question. \\
Recent neurobiological breakthroughs \cite{larkum1999new, larkum2013cellular, phillips2017cognitive} have revealed
that two-point layer 5 pyramidal cells
(L5PCs) in the mammalian neocortex use their apical inputs as context to modulate the transmission of coherent feedforward (FF) inputs to their basal dendrites (Figure 1) \cite{larkum1999new, larkum2013cellular, major2013active, ramaswamy2015anatomy, phillips2017cognitive, larkum2022dendrites, adeel2020conscious, kording2000learning, SchumanAnnual, poirazi2020, larkum2018perspective}. Such modulatory regulation via apical dendrites has been associated with the flexibility and reliability of neocortical dynamics \cite{shine2016dynamics, shine2019human, shine2019neuromodulatory}. For example, a rigorous dynamic systems perspective \cite{shine2021computational} suggests that neuromodulation selectively upregulates, and thus flexibly integrates, a subset of disparate cortical regions that would otherwise operate more independently. At a granular level, a recently reported dataset directly recorded from slices of rodent neocortex shows how L5PCs process information in a context-sensitive manner \cite{schulz2021gaba, kay2020contextual, kay2022comparison} e.g., the L5PC transmits unique information about the FF data without transmitting any unique information about the context. However, depending upon the strength of the FF input, the context adds synergy, which is the information requiring both the context and FF input. These studies test the relationship between context and FF inputs and convincingly validate the biological plausibility of the context-sensitive style of neural information
processing.
% \begin{wrapfigure}{r}{0.75\textwidth}
% 	\centering
% 	\includegraphics[trim=0cm 0cm 0cm 0cm, clip=true, width=0.5\textwidth]{sciencetry1.pdf}
% 	\caption{Context-sensitive neocortical neuron whose apical dendrites are in layer 1 (L1) with cell body and basal dendrites in deeper layers. The apical tuft receives input from diverse cortical and subcortical sources as context to amplify and suppress the transmission of coherent and incoherent FF signals, respectively. However, the kinds of information that arrive at the apical tuft and how they influence the cell’s response to the feedforward (FF) input remain unclear. Therefore, these neurons have not been widely exploited by state-of-the-art DL models.}
% 	\label{l5pc}
%  \vspace{-1em}
% \end{wrapfigure}
% where action potentials (APs) suppress when the FF input is low and amplify when: (i) FF input is high, (ii) FF input is low and context input is high, or (iii) both FF and context inputs are high; 
% inputs to the apical dendrites have distinct effects on the output that context sensitivity implies e.g., the L5PC transmits unique information about the FF data without transmitting any unique information about the context. However, depending upon the strength of the FF input, the context adds synergy, which is the information requiring both the context and FF input. 
% \cite{schulz2021gaba}\cite{kay2020contextual}\cite{kay2022comparison}
% where 
% These computational studies are the first to test the relationship between context and FF inputs using real rats' somatosensory cortical L5PC data and 
\\Despite rapidly growing neurobiological evidence suggesting that context-sensitive two-point neurons are fundamental for optimal learning and processing in the brain and could circumvent the computational limitations of DL, the computational potential of these neurons to process large-scale complex real-world data remained underestimated \cite{gidon2020dendritic ,larkum2022dendrites, sarwat2022chalcogenide}. Therefore, these neurons have not been widely exploited by state-of-the-art DL models. Although a few machine learning studies such as \cite{ payeur2021burst, sacramento2018dendritic, guerguiev2017towards} have been inspired by the discovery of two-point neurons, these methods focused predominantly on using apical inputs for credit assignment (learning). In contrast, the apical input from the feedback and lateral connections is multifaceted and far more diverse with far greater implications for ongoing learning and processing in the brain than realised \cite{adeel2020conscious}. Therefore, to fully benefit from the capabilities these neurons have to offer, it is critical to understand the kinds of information that arrive at the apical tuft and their influence on the cell’s response to the FF input. Inspired by the latest fundamental
advances in cellular neurobiology \cite{larkum1999new, larkum2013cellular, phillips2017cognitive, major2013active, ramaswamy2015anatomy, shine2016dynamics, shine2019human, shine2019neuromodulatory, shine2021computational, kording2000learning, SchumanAnnual, poirazi2020, larkum2018perspective, schulz2021gaba, kay2020contextual, kay2022comparison, gidon2020dendritic, aru2020cellular, bachmann2020dendritic, shin2021memories, benjamin2021neocortical, heeger2017theory, heeger2019oscillatory, heeger2020recurrent}, here we address these issues and demonstrate that context-sensitive two-point neurons have information processing capabilities of the kind displayed by the neocortex and can circumvent the computational limitations of DL.

\textbf{Results}
\begin{figure*} 
	\centering
	\includegraphics[trim=0cm 0cm 0cm 0cm, clip=true, width=1\textwidth]{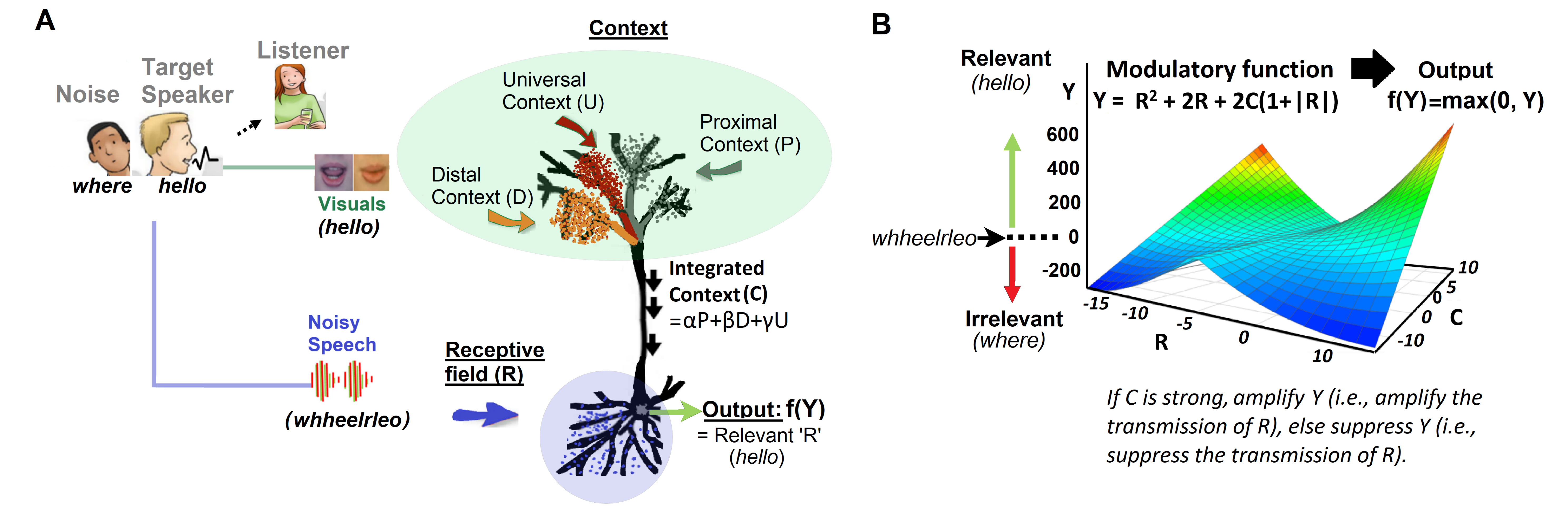}
	\caption{\textbf{Context-sensitive two-point neuron inspired-cooperative context-sensitive neural information processing applied to audiovisual speech denoising in a challenging multi-talker environment.} \textbf{(A)} Schematic diagram of the two-point neuron-inspired cooperative context-sensitive auditory processor that receives input from diverse sources as a context to amplify and attenuate the transmission of relevant and irrelevant FF information received at the basal, respectively. The processor receives three different kinds of context, proximal (P), distal (D), and universal (U): P and D represent information from the neighbouring auditory processors and distal visual processors, respectively, and U represents cross-modal memory (Figure S1 (A-B)). However, U could explicitly be extended to the sources of inputs to include prior experiences, emotional states, and cognitive load. The integrated context (C) via modulatory function amplifies and suppresses the transmission of relevant and irrelevant speech signals heard in noisy environments, where $\alpha$, $\beta$, and $\gamma$ are the weights associated with P, D, and U, respectively. \textbf{(B)} The modulatory function uses C as a driving force to split the signal into relevant and irrelevant signals. It amplifies the output when C is high and suppresses the output when C is low. The Rectified Linear Unit (ReLU) discards the suppressed information (below zero). The context-sensitive deep information processing architecture composed of context-sensitive processors (Figure S1 (C)), turns off up to 99\% of units carrying irrelevant information. As opposed to Infomax, which maximises the transmission of information irrespective of whether or not it is relevant in the current context, the proposed approach maximises the transmission of information that is relevant in the current context. This distinction is at the core of the proposed approach and is not just sparse coding.} 
	\label{l5pc}
 \vspace{-1em}
\end{figure*}

Figure 2 illustrates a context-sensitive two-point neuron-inspired  cooperative context-sensitive neural information processing mechanism applied to robustly deal with speech-in-noise (SIN) \cite{mcgurk1976hearing, middelweerd1987effect, macleod1990procedure, adeel2019lip, adeel2020contextual}. Specifically, Figure 2A depicts a single context-sensitive two-point auditory processor that receives input from diverse sources at the apical and uses it as context to amplify and suppress the transmission of relevant and irrelevant FF speech signals received at the basal, respectively. For example, the processor uses information from distal visual processors as distal context (D), neighbouring auditory processors as proximal context (P), and cross-modal working memory (M) as universal context (U) (see Figure S1 for detailed information flow and the formation of contextual fields). The context-sensitive processor uses integrated context (C) via asynchronous modulatory transfer function (AMTF) (Figure 2B) to selectively amplify and suppress the FF transmission of the relevant and irrelevant auditory information, respectively.\\
The proposed AMTF uses C as a driving force to split the signal into relevant and irrelevant signals. In previously proposed AMTFs (Figure S2) \cite{kay2020contextual, kay2022comparison}, R drives the firing of two-compartment L5PC. If R is strong, context is neither necessary nor sufficient for the neuron to transmit information about the R. If R is very weak (or does not exist), even a very strong context does not encourage firing. Here we show that context can overrule the strength of the R and can conversely discourage or encourage firing if R is strong or weak, respectively. This new AMTF uses context as a `modulatory force' to push the processor output to the positive side of the activation function (e.g., ReLU) if R is important, otherwise to the negative side. This mechanism enhances cooperation and seeks to maximise agreement between the active processors. Nonetheless, the modulatory force that enables this move systematically could be generated in several different ways, linearly or non-linearly e.g., instead of ReLU, half-Gaussian filter could be used. The modulatory transfer function could be seen as a signalling module that signals `Yes' with certain confidence if a match between data streams has been found regarding a specific sensory or cognitive feature. \\
Figures S1 (C), S2, and S3 depict example context-sensitive processors-driven deep convolutional neural net architectures. Here conventional point processors are used to generate R, P, D, and U, whereas context-sensitive processors are used in non-parametric modulatory (NPM) blocks for selective audiovisual (AV) information processing. Each layer conditionally segregates the relevant and irrelevant information streams, and then recombines only the relevant streams to extract cross-modal brief memory \cite{adeel2020conscious, aru2020cellular, bachmann2020dendritic, shin2021memories, benjamin2021neocortical}, which is broadcasted and received by processors with the current P and D in the next layer. Here the brief working memory could be seen as if the selected relevant receptive fields (Rs) are temporarily preserved at time t-1, while attention at time t is engaged with the upcoming R e.g., holding a person's address in mind while listening to instructions about how to get there. This is the ability of the network to retain information for a short period of time \cite{poirazi2020}. In general, U could explicitly be extended to the sources of inputs to include general information about the target domain acquired from prior experiences, emotional states, intentions, cognitive load, and semantic knowledge. The contextual fields P, D, and U could be calculated in several different ways (see Figure S3, Table S1 and Table S2).\\
This basic context-sensitive neural information mechanism includes many of the anatomical and functional elements observed in slices of rodent neocortex. While our model is extremely simplified, it captures critical processing steps found, e.g., in \cite{larkum1999new, larkum2013cellular, phillips2017cognitive, schulz2021gaba, kay2020contextual, kay2022comparison, gidon2020dendritic} where the apical input amplifies and suppresses the transmission of FF input; in \cite{aru2020cellular, bachmann2020dendritic, shin2021memories, benjamin2021neocortical} where  the apical tuft incorporates input from both thalamic and different cortical sources to enable conditional segregation and recombination of multiple input streams; in the dataset \cite{schulz2021gaba, kay2020contextual, kay2022comparison, gidon2020dendritic} recorded from slices of rodent neocortex that shows how inputs to the apical dendrites have distinct effects on the output that context sensitivity implies; in \cite{baars2005global, de2015nonspecific} where the modulatory effect of brief memory formation and retrieval is broadcasted to all sensory modalities; and in \cite{heeger2017theory, heeger2019oscillatory, heeger2020recurrent} where apical amplification is described as recurrent drive and the recurrent weights are similar to the contextual fields specified by synapses of the apical dendrites. Furthermore, how such functional models can simulate key sequential phenomena of working memory, and in addition to being temporarily stored, information can be modified by complex sequential dynamics \cite{heeger2020recurrent}.\\ 
The proposed mechanism is compared against the point processors-driven variational autoencoder (VAE) and the vanilla version of our model (baseline). For a fair comparison, the baseline/ state-of-the-art convolutional deep model is integrated with the cross-channel communication (C3)/attention blocks \cite{yang2019cross, cangea2019xflow, guo2019deep, bhatti2021attentive}. The baseline models implement C3 or cross-channel fusion through concatenation, addition, or multiplication using the `point' processor \cite{burkitt2006review}. Thus, each processor integrates all the incoming streams in an identical way i.e., simply summing up all the excitatory and inhibitory inputs with an assumption that they have the same chance of affecting the processor's output \cite{hausser2001synaptic}. The deep models reconstruct an ideal binary mask (IBM) \cite{gogate2018dnn} or clean short-time Fourier transform (STFT) of the audio signal given noisy audio and visuals \cite{adeel2020contextual, gogate2020cochleanet}. All deep models are trained using the benchmark AV Grid \cite{cooke2006audio} and ChiME3 \cite{barker2017third} corpora, with 4 different real-world noise types (cafe, street junction, public transport (BUS), pedestrian area) (see materials and methods for more details and Table S3). Comparative results demonstrate that context-sensitive processors can process large amounts of AV information effectively and efficiently compared to point processors.
\begin{figure*} 
	\centering
	\includegraphics[trim=0cm 0cm 0cm 0cm, clip=true, width=1\textwidth]{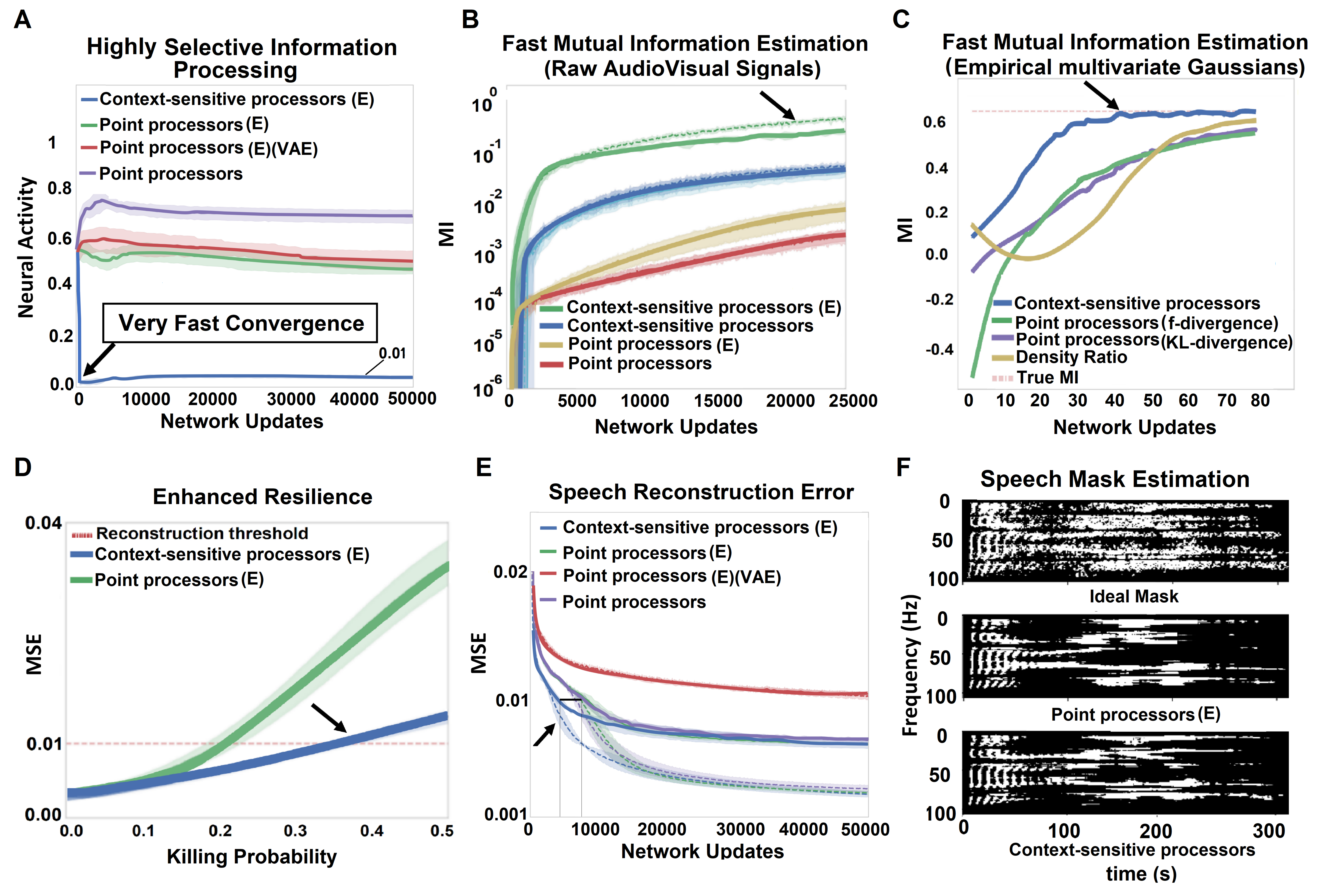}
	\caption{\textbf{Context-sensitive processors can efficiently process large amounts of heterogeneous real-world AV data.} \textbf{(A)} Selective information processing: the blue line shows that context-sensitive processors quickly evolve to become highly sensitive to relevant information and become active only when the received information is important for the task at hand. In contrast, point processors-driven baseline model and $\beta-$variational autoencoder (VAE) with and without energy term (E) in the cost function experience significantly higher neural activity. \textbf{(B)} Mutual information (MI) estimation and maximization between high dimensional clean visual and noisy speech signal. Note that the context-sensitive processors-driven deep model converges quickly to the higher MI. The negative MI is due to untrained random weights at the start of the neural net training. Solid and dashed lines indicate testing loss and training phases, respectively. \textbf{(C)} To test the system against true MI, the network is used to estimate and maximize MI between multivariate Gaussian Random Variables. It can be observed that context-sensitive processors quickly converge to the true MI compared to other sophisticated point-processor driven methods, including MI neural estimation with \textit{f} and Kullback–Leibler (KL) divergence \cite{belghazi2018mutual}. \textbf{(D)} Resilience test: when trained models were tested for resilience with 35\% randomly killed processors, context-sensitive processors degraded performance gracefully as compared to point processors. \textbf{(E-F)} AV speech reconstruction error and speech mask estimation: the context-sensitive processors-driven deep model achieves comparable results with faster learning at the early training stage despite using significantly less number of processors at any moment.} 
	\label{l5pc}
 \vspace{-1em}
\end{figure*}
\begin{figure*} 
	\centering
	\includegraphics[trim=0cm 0cm 0cm 0cm, clip=true, width=1\textwidth]{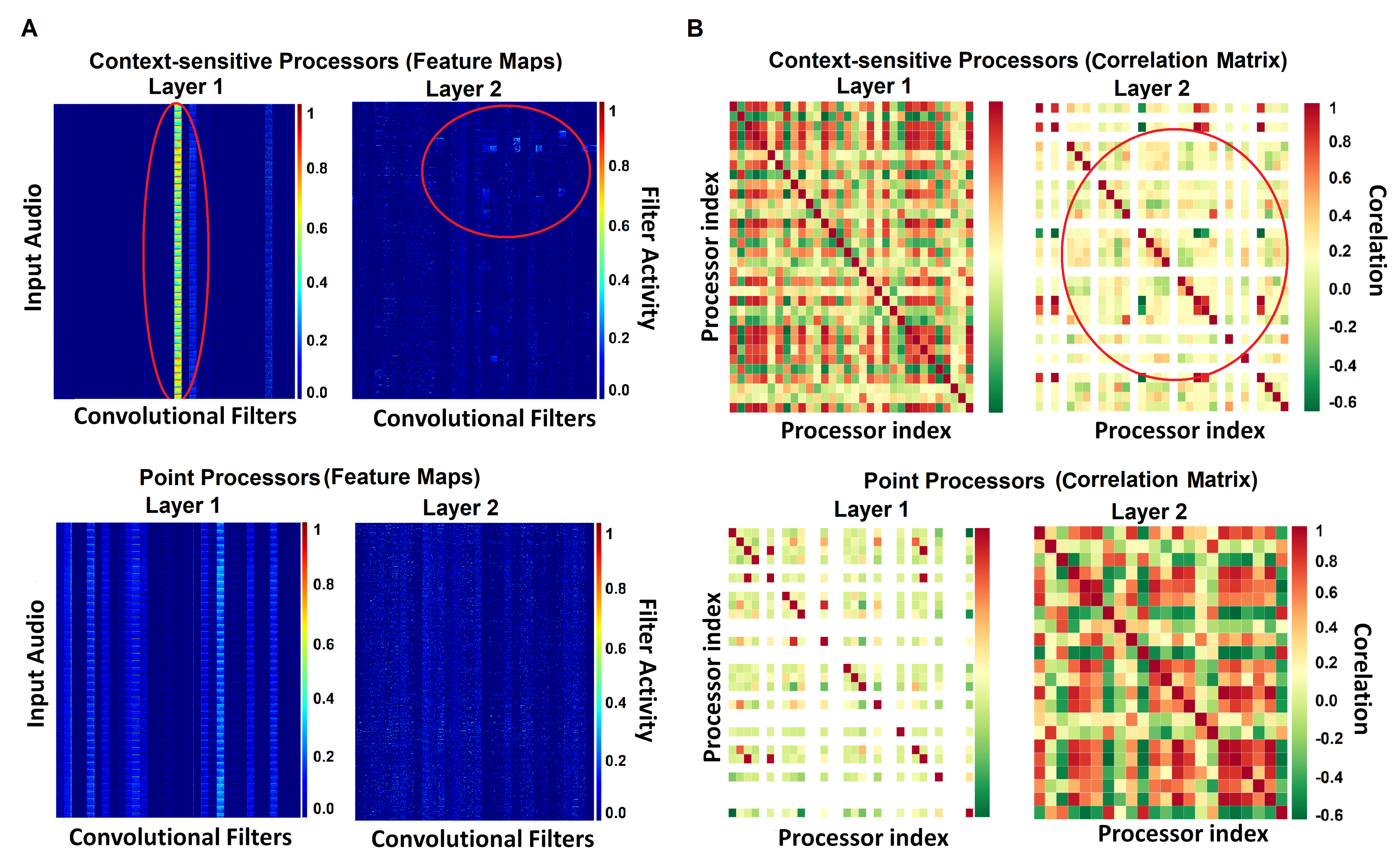}
	\caption{\textbf{Context-sensitive local processors transmit only relevant information:} \textbf{(A)} Feature maps: the Y-axis represents the input speech signal of 240ms duration, where
each small block is of 10ms duration. The X-axis represents 32 convolutional filters. It is to be observed that context-sensitive processors are able to effectively amplify and suppress the transmission of relevant and irrelevant signals, respectively. For example, here low-level layers are restricting the transmission of irrelevant information to higher levels i.e., far fewer  filters in Layer 1 and Layer 2 are active compared to point processors-driven deep feature maps. In addition, it is to be noted that context-sensitive processors could construct high-level representation of the output at low-level layers requiring less number of processors to construct a good representation. \textbf{(B)} The data from 32 processors show that context-sensitive processors reduce the cross-correlation as the data passes through different layers compared to the point processors.}
 \vspace{-1.2em}
\end{figure*}
\\\textbf{Context-sensitive processors guide audio signal processing towards the currently relevant information far more effectively and efficiently than current forms of DL:} For AV speech denoising on Grid and ChiME3 datasets, we used a deep net comprising an input layer, two Convolution layers, and an output layer. Audio and visual features of time instances $t_k, t_{k-1},....t_{k-5}$ were fed into the model, where \textit{k} represents the time instance. See methods and supplementary material for detailed configurations and parameters. Training results demonstrate that a context-sensitive processors-driven deep net can reconstruct clean speech using far less number of processors compared to conventional point processors-driven deep net. Figure 3A depicts selective information processing results. It is to be observed that context-sensitive processors quickly evolve to become highly sensitive to a specific type of high-level information and `turn on' only when the received signals are relevant in the current context. This allows the network to be selective as to what data is worth paying attention to and therefore processing that, instead of having to process everything. This reduction in neural activity is equivalent to a magnitude of energy efficiency during training if the synapses associated with the cells with zero activity are turned off in the hardware. Overall, in the network of 16 million parameters, the context-sensitive processors reduce their neural activity to 0.01 compared to the baseline, which converges to the neural activity of 0.45. Remarkably, context-sensitive processors achieve this low activity in just a few training updates. For a larger model comprising 40 million parameters, the context-sensitive processors reduce their activity to less than 0.008\% i.e., ~1250x less (per FF transmission) than the baseline (Figure S4). However, the reconstruction accuracy for both point and two-point processors drop. In this case, more tuning and optimisation are required to search for Pareto-optimal. When the context-sensitive processors are trained without memory, they converge to the overall neural activity of ~0.2 (Figure S5). This suggests that selective information processing is highly dependent on the strength of context.\\
The effect of selective information processing is evident in mutual information (MI) estimation between high dimensional clean visual and noisy speech signals (Figure 3B). It is to be observed that the baseline models remain deficient in achieving high MI, regardless of the experimental setup, hyper-parameters, and loss function. In contrast, context-sensitive processors driven deep model converges quickly to the higher MI. We also remark that context-sensitive processors converge quickly to the true MI when the same network is used for multivariate Gaussian random variables \cite{belghazi2018mutual} and compared against three popular point processors-driven MI neural estimation methods: f-divergence, KL-divergence, and density ratio \cite{belghazi2018mutual} (Figure 3C). 
Furthermore, context-sensitive processors are inherently robust against sudden damage. It is to be observed that when trained deep models
 are tested for resilience with up to 35\% randomly killed processors, context-sensitive processors degrade performance gracefully compared to point processors (Figure 3D). Despite using a few processors at any moment, context-sensitive processors-driven deep net enable faster learning at the early training stage (Figure 3E) with comparable prediction accuracy (Figure 3F).\\
Similar results achieved when deep models were used to reconstruct high dimensional short-time Fourier transform (STFTs) of the clean speech (Figures S6-S7). In this case, the quality of reconstruction with context-sensitive processors remained sensitive to fine-grained details and distinguished the relevant signal more easily and clearly. This had a significant impact on the reconstructed time-domain speech signal and its intelligibility (Figure S8) \cite{adeel2020contextual}. We conclude that context-sensitive processors-driven deep net can effectively process large amounts of heterogeneous real-world AV data using far fewer processing units at any moment than point-processors-driven deep net. \\
Figure 4A reveals the selective amplification and suppression properties of context-sensitive processors compared to the point processors. It is to be observed that the baseline treats each input with equal importance and computes features regardless of the underlying nature of the signal. On the contrary, context-sensitive processors highlight relevant and irrelevant features. This analysis could also be seen as a Fourier analysis or time-frequency analysis explaining what matters when. We hypothesise that context-sensitive processors highlight which phonemes matters the most and discover the aspects of speech seen in the video or the structure (high-level features) at early layers. These results also show that context-sensitive processors have more processing  ability to make important decisions at the cellular level. In general, these patterns are certainly providing important information as compared to the baseline. We found similar behaviour for visual features (Figure S9) and audio features in different SNRs (Figure S10). Furthermore, Figure 4B provides further insight by depicting how the data is statistically transformed through different deep layers. The autocorrelation and cross-correlation data from 32 processors are shown. It is to be observed that cross-correlation reduces significantly in the case for context-sensitive processors when data moves from one layer to the next i.e., more information passes on to the next block. In contrast, the baseline passes more redundant data to the next layer as high cross-correlation could be observed.
\textbf{Discussion}
\\Results support our hypothesis that the fundamental weakness of state-of-the-art deep learning is its dependence on a long-established simplified point processor that maximises the transmission of information regardless of its relevance to other processors or the long-term benefit of the whole network. In contrast, context-sensitive processors cooperate moment-by-moment and transmit information only when the received FF information is coherent to the overall activity of the network or relevant to the task at hand. This new style of cooperative context-sensitive neural information processing enables relevant feature extraction at very early stages in the network, leading to faster learning, reduced neural activity, and enhanced  resilience. 
\\Although point processors allow DL to learn the representation of information with multiple levels of abstraction, their processing is shallow \cite{craik1972levels}. Specifically, point processors encode the FF information based on repetition activity (learning) without any search for coherence. In contrast, our proposed cooperative context-sensitive neural information processing promotes deep information processing (DIP) \cite{craik1972levels} that allows individual processors to have more deeper and well-reasoned interaction with the received FF information. For example, the demonstration of relevant and irrelevant signals amplification and suppression, respectively, and the construction of high-level features at low-level layers show how low-level layers can make strategic decisions and restrict the transmission of conflicting information to the higher layers to avoid disorganization and achieve harmony. \\
It is worth mentioning that our work is not a model, but a demonstration that the cooperative context-sensitive  style of computing has exceptional big data information processing capabilities. Our contribution to this rapidly growing field of research encourages machine learning experts to exploit context-sensitive processors in state-of-the-art DL models for applications where speed and the efficient use of energy are crucial. It also encourages neurobiologists to search for the essentials (fine-tunings) which were necessary to make this neurobiological mechanism work. \\
We learnt that context plays an essential role in selective information processing. Specifically, when the processors process noisy information, context overrules the typical dominance of the receptive field and, therefore, drives neural activity. Furthermore, the higher the context, the higher the efficient information processing. For example, when train without memory (universal context), context-sensitive processors reduce the overall neural net activity but far less than processors with memory (Figure S4). We also found that memory components could be effectively formed through conditional segregation of relevant and irrelevant information streams e.g., when conditional segregation and recombination of multiple input streams are transformed into memory, along with other contextual fields, it improves selective amplification and suppression of relevant and irrelevant signals, respectively. Therefore, the formation of different kinds of contexts arriving at the apical and their influence on the cell’s response to the FF input are crucial for  context-sensitive neural information processing. \\
% Although a few recent machine learning models such as \cite{lillicrap2020backpropagation, guerguiev2017towards, payeur2021burst, sacramento2018dendritic} have been inspired by the role of apical tuft in L5PC, they do not interpret them as evidence for context-sensitive two-point neurons as we do here. Specifically, the earlier networks focused on learning, whereas our model predominantly focused on using context to guide ongoing processing. Therefore, we are able to develop new forms of context-sensitive information processing mechanisms that can outperform the best available forms of deep learning inspired by the `point’ neuron assumption. \\
The proposed work also bridges the gap between Dendritic Integration Theory \cite{bachmann2020dendritic} and Global Neuronal Workspace Theory \cite{baars2005global} e.g., the notion of universal context matches with the universal role of the NSP-thalamus whose modulatory effect is broadcasted to all sensory modalities activating other brain areas \cite{de2015nonspecific}. We suggest that in addition to the synergy between apical and basal information flow in L5PC \cite{aru2020cellular}, the synergy between different coherent information streams could be closely related to the L5PC processing. Whether this holds true or not, the role of universal context is of great importance since things are experienced differently in positive and negative frames of mind and with different intentions, attentions, hopes, and emotional states. We suggest that the universal context may be analogous to signals that regulate the balance between apical/internal/top-down/feedback and basal/external/sensory/FF inputs. Thus, switching the mode of apical function between amplification (or drive) and isolation. If so, the idea of the universal context may be relevant to a major physiological process that is only now being seen to be important \cite{marvan2021apical, shin2021memories}. Finally, the notion of universal context leads us to think that the `self' is an enduring part of the internal context. So, are `we' the enduring internal context within which our experiences occur? \\
Our work supports the argument that the leaky integrate-and-fire conception of the neuron harms our progress in understanding brain function \cite{larkum2022dendrites}. Therefore, the proposed work could help to better understand neurodevelopmental disorders such as autism and sensory overload when early brain layers fail to filter out irrelevant information and the brain becomes overwhelmed due to excessive contradictory messages transmitted to higher perceptual levels \cite{rinaldi2008hyper, markram2010intense}, or epilepsy when the bursts of electrical activity in the brain cause seizures. Last but not least, the proposed work sheds light on human's basic cooperative instinct that achieves harmony  via organized cooperation between diverse neurons \cite{ridley1996origins, rilling2002neural, delle2016lay}. For example, the review evidence shows unequivocally that changes in brain state such as those from sleeping to waking or from low to high arousal depend on the neuromodulatory regulation of apical function in pyramidal cells \cite{tantirigama2020perspective}. It is shown how impairments of apical dendritic function have a key role in some common neurodevelopmental disorders, including autism spectrum disorders \cite{nelson2021dendritic}. The apical dendritic mechanisms rooted in genetic foundations experience specific genetic mutations that impair these fundamental cellular mechanisms. A few convincing reviews \cite{aru2019coupling, aru2020cellular, shepherd2021untangling, marvan2021apical} also suggest that the thalamocortical loops with a key role in conscious experience depend on apical dendrites in L1. \\
Overall, to the best of our knowledge, this is the first time context-sensitive two-point L5PC mechanism has been applied to solve any challenging real-world problem, reflecting its potential to transform the capabilities of neurocomputational systems. We believe that the proposed cooperative context-sensitive style of information processing, supported by the latest and rapidly growing neurobiological discoveries on two-point cells, may be fundamental to the capabilities of the mammalian neocortex. The context sensitivity at the cellular level indeed has information processing abilities of the kind displayed by the mammalian neocortex. \\
Ongoing work involves using local context as a feedback error e.g., for credit assignment, as opposed to the way it is typically used for training standard deep learning algorithms \cite{guerguiev2017towards, payeur2021burst}. We aim to provide further insights into cooperative context-sensitive learning mechanism. Ongoing work also involves the demonstration of the proposed modulatory concept within unimodal streams to extend cooperative context-sensitive information processing well beyond multimodal applications. Although our results demonstrated how video modulates the transmission of auditory information and vice versa, the clean video available at each of the deep layers in our architecture may be guiding the discovery of structure shared by the audio and visual streams. Thus, ongoing work includes analysis of the trivariate MI components transmitted by each of the deep layers in our architecture that we believe may provide a wholly new perspective on the multisensory processing and `merging of the  senses’ in neocortex that has long been studied by many neurobiologists and psychologists. \\
\textbf{Materials and Methods}\\
\textbf{Context-sensitive processor:}\\
For the sake of mathematical simplicity and generality across this section, we use Einstein tensor notation. We also reduce the discussion to vector spaces indexed by a single element as in machine learning we are only interested in numerable collections of vector spaces. Nonetheless, in some cases, it may be useful to include certain topological properties as different indices i.e. an image can be represented as $\tensor{Z}{}{}{ \alpha \beta \gamma}$. In addition, we restrict ourselves to the simple case of two channels and denote the analogue variable for the other channel with a bar and, in a huge abuse of notation, we denote every learnable variable with $\tensor{\boldsymbol\theta}{}{}{}$. Unlike previous works, our idea is to compute only the relevant information shared between channels while, at the same time, preventing local non-important information from each channel to overtake the computation. Thus, we consider a family $\mathcal{F}$ of parametric functions $f$ composed almost entirely by transformations $h: V_{\alpha} \times V_{\beta} \mapsto V_{\gamma}$.
\begin{align}
    R: \tensor{r}{ \ell }{}{ \eta }
    &= \tensor{\boldsymbol\theta}{ \ell }{ \alpha }{ \eta } \tensor{A}{ \ell - 1 }{}{ \alpha }
\end{align}   
\begin{align}
    P: \tensor{p}{ \ell }{}{ \mu }
    &= \tensor{\boldsymbol\theta}{ \ell }{ \eta }{ \mu } \tensor{r}{ \ell }{}{ \eta }
\end{align}   
\begin{align}
    D: \tensor{d}{ \ell }{}{ \nu }
    &= \tensor{\boldsymbol\theta}{ \ell }{ \tau }{ \nu } \tensor{\bar{r}}{ \ell - 1 }{}{ \tau }
\end{align}   
\begin{align}
    U: \tensor{m}{ \ell - 1 }{}{ \xi }
    &= \tensor{\boldsymbol\theta}{ \ell }{ \rho }{ \xi } \tensor{m}{ \ell - 2 }{}{ \rho } + \tensor{\boldsymbol\theta}{ \ell }{ \alpha }{ \xi } \tensor{A}{\ell - 1}{}{ \alpha } \nonumber \\
    &\phantom{=\tensor{\boldsymbol\theta}{ \ell }{ \rho }{ \xi } \tensor{m}{ \ell - 2 }{}{ \rho }~} + \tensor{\boldsymbol\theta}{ \ell }{ \beta }{ \xi } \tensor{\bar{A}}{\ell - 1}{}{ \beta }
\end{align}   
\begin{align}
    C: \tensor{C}{ \ell }{}{ \epsilon }
    &= \tensor{ \boldsymbol\theta }{ \ell }{ \mu \nu \xi }{\epsilon } \tensor{P}{ \ell }{}{ \mu } \tensor{D}{ \ell }{}{ \nu } \tensor{U}{ \ell - 1}{}{ \xi }
\end{align}  
\begin{align}
    \tensor{a}{ \ell }{}{ \gamma }
    &= \tensor{ \Delta }{}{ \eta \epsilon }{ \gamma } \tensor{r}{ \ell }{}{ \eta } \tensor{C}{ \ell }{}{ \epsilon }
\end{align}  
\begin{align}
    h \parentheses{\tensor{A}{\ell - 1}{}{ \alpha }, \tensor{\bar{A}}{\ell - 1}{}{ \beta }; \tensor{\boldsymbol\Theta}{}{}{}} :=
    \tensor{A}{ \ell }{}{ \gamma }
    &= \zeta\parentheses{ \tensor{a}{ \ell }{}{ \gamma } }
\end{align}  
where $\tensor{\boldsymbol\Theta}{}{}{} = \{ \tensor{\boldsymbol\theta}{ \ell }{ \alpha }{ \eta }, \tensor{\boldsymbol\theta}{ \ell }{ \eta }{ \mu }, \tensor{\boldsymbol\theta}{ \ell }{ \tau }{ \nu }, \tensor{\boldsymbol\theta}{ \ell }{ \rho }{ \xi }, \tensor{\boldsymbol\theta}{}{ \alpha }{ \xi }, \tensor{\boldsymbol\theta}{ \ell }{ \beta }{ \xi }, \tensor{ \boldsymbol\theta }{ \ell }{ \mu \nu \xi }{\epsilon } \}$ is the collection of learnable parametric linear transformations of $h$; the operator $\tensor{ \Delta }{}{ \eta \epsilon }{ \tau }$ denotes the hadamard product between $\tensor{r}{ \ell }{}{ \eta }$ and $\tensor{c}{ \ell }{}{ \epsilon }$. Notice that this implicitly assumes that the vector space of both operands is of the same size, and $\zeta$ is the activation function. In practice, we also consider another set of trainable variables $\tensor{\boldsymbol\lambda}{\ell}{}{ \kappa }$ which are added to the result of each transformation but including them in the previous equations may obscure the most relevant part of the computation. We can replace the operator $\tensor{ \Delta }{}{ \eta \epsilon }{ \tau }$ with other operators to simulate a more complex relationship between R and C. We suspect that the exploration of better modulatory operators may play a major role in the near future. Intuitively, we enforce variables $\tensor{p}{\ell}{}{\nu}$ and $\tensor{d}{\ell}{}{\mu}$ to extract the core information that is currently held in the other parallel streams. Similarly, we enforce the term $\tensor{m}{\ell}{}{ \xi }$ to act as a collective reservoir of important information extracted at a previous layer from both channels.\\
For MI estimation and speech denoising we used the following loss functions:
\begin{align*}
    \mathcal{L}_{1} &= -\alpha\expectation{-I_{f}\parentheses{\tensor{X}{}{}{} ; \tensor{Y}{}{}{} }}{} + \gamma\expectation{\mathcal{E}}{} \\
    \mathcal{L}_{2} &= \beta \expectation{\text{SE}\parentheses{\tensor{Z}{}{}{},\tensor{\hat{Z}}{}{}{}}}{} + \gamma\expectation{\mathcal{E}}{}
\end{align*}
$\mathcal{E}$ is a differentiable approximation for the number of firings. 
We adjust the coefficients of the loss functions to make the secondary objectives significantly less important than the main goal; in particular, we set $\gamma$ to a really small value in all experiments. Even for very small $\gamma$, we encounter that the gradient signal from the energy may be several orders of magnitude greater than the signal originated from the MI estimation.
\\
\textbf{MI estimation:} Multimodal (MM) representation learning via MI maximization has proven to significantly improve both the classification and regression tasks \cite{belghazi2018mutual,velickovic2019deep,liao2021multimodal}. However, MI maximization between high dimensional input variables in the presence of extreme noise is a serious challenge. Here we pose a problem of learning MM representation via estimating and maximizing the MI between high dimensional clean visual and noisy speech signals. For  direct computation of the entropy or the Kullback–Leibler divergence, we uses Donsker-Varadhan representation. In other words, we transformed the mutual information estimation problem into an optimization problem \cite{belghazi2018mutual}.\\
Consider $\tensor{X}{}{}{\alpha} \in V_{\alpha}$  and $\tensor{Y}{}{}{\beta} \in V_{\beta}$ two random multidimensional variables indexed by $\alpha \in \{1, \dots, A\}, \beta \in \{1, \dots, B\}$, with $V_{\alpha} \subseteq \mathbb{R}^{A}$ and $V_{\beta} \subseteq \mathbb{R}^{B}$ and distributed as $\mathbb{P}_{X}$ and $\mathbb{P}_{Y}$, respectively.\\
The mutual information between these two variables, $I\parentheses{\tensor{X}{}{}{\alpha} ; \tensor{Y}{}{}{\beta}}$, is given by,
\begin{align}
    I\parentheses{\tensor{X}{}{}{\alpha} ; \tensor{Y}{}{}{\beta}}
    &= H\parentheses{\tensor{X}{}{}{\alpha}} - H\parentheses{\tensor{X}{}{}{\alpha} \mid \tensor{Y}{}{}{\beta}} \nonumber 
    &= \mathcal{D}_{KL} \parentheses{ \mathbb{P}_{X} \parallel \mathbb{P}_{Y} }
\end{align}
In general, direct computation of the entropy or the KL divergence is not feasible. Fortunately, it is possible to rewrite this expression using the Donsker-Varadhan representation. Thus,

\begin{align}
    I\parentheses{\tensor{X}{}{}{\alpha} ; \tensor{Y}{}{}{\beta} }  
    &= \sup_{f \in \mathcal{F}} I_{f}\parentheses{\tensor{X}{}{}{\alpha} ; \tensor{Y}{}{}{\beta} } \nonumber \\
    &= \sup_{f \in \mathcal{F}} \expectation{f\parentheses{\tensor{X}{}{}{\alpha}, \tensor{Y}{}{}{\beta}}}{\mathbb{P}_{XY}} \nonumber \\
    &\hspace{-0.9cm}- \log \parentheses{ \expectation{\exp \parentheses{ {f\parentheses{\tensor{X}{}{}{\alpha}, \tensor{Y}{}{}{\beta}}}}}{\mathbb{P}_{X} \times \mathbb{P}_{Y}} }
\end{align}
where $\mathcal{F}$ is a set of functions $f: V_{\alpha} \times V_{\beta} \mapsto \mathbb{R}$ with finite expectations under $\mathbb{P}_{XY}$ and $\mathbb{P}_{X} \times \mathbb{P}_{Y}$.

Hence, $\forall f \in \mathcal{F}$ we have:
\begin{align}
    I\parentheses{\tensor{X}{}{}{\alpha} ; \tensor{Y}{}{}{\beta} } \geq I_{f}\parentheses{\tensor{X}{}{}{\alpha} ; \tensor{Y}{}{}{\beta}}
\end{align}\\
\textbf{AV corpus:} For AV speech processing, the Grid \cite{cooke2006audio} and ChiME3 \cite{barker2017third} corpora are used \cite{adeel2020contextual}, including four different noise types; cafe, street junction, public transport (bus), and pedestrian area with the signal-to-noise ratio (SNRs) ranging from -12dB to 12dB with a step size of three. Grid and ChiME3 corpora are publicly available and open-source, thus, ethical approval is not needed. See Table S3 and Figure S11.\\
\textbf{Deep multimodal supervised reconstruction:} For this task, we used the mask estimation approach for speech enhancement presented in \cite{gogate2018dnn, gogate2020cochleanet}. See supplementary material for more details.
\\\textbf{Simulation details:} All deep models have a similar structure, layers, and configuration. We used two convolutional layers, each with 32 filters, kernels of size 5 and stride 2. For each channel embedding, we used 128 units and for the global embedding, we used 256 units. Additional terms to the losses, like ELBO loss, were added to the loss function model-wise. All activation functions are ReLUs. All networks are initialized with a glorot uniform distribution. The Adam optimizer with a learning rate of $1e^{-6}$ and $1e^{-4}$ is used for all the experiments. Although we do not claim these configurations are optimal, we empirically observed models behaved well with this set of parameters.\\
Each element of the dataset is a tuple containing a noisy audio signal (ideal binary mask (IBM) \cite{gogate2018dnn}, STFT), a snapshot of the lips of the speaker (image), and a clean audio signal. The SNR varies from +12dB to -12dB in steps of 3dB. The noisy audio was corrupted with several different noise sources. Although, more sophisticated approaches for denoising using neural networks exist, our goal is to measure the capabilities of the network using as few resources as possible (neural activity). For this experiment, we introduced a small change into the modulatory step, in which we also take the activity of neighbouring processors into account twice, once when we compute the context and again when we apply the modulation. This small change is equivalent to replacing the delta operator of equation 6. 
As usual in machine learning, we take a split 80\%-20\% for training and testing; we leave a single sample out of training and testing splits to use as a proxy for the figures in this work.
Data is normalized across the whole dataset and presorted to break all order correlations. 
The dataset is shuffled once more with a seed to add some variability between different runs and to ensure that different models encounter a similar landscape.
We use a mini-batch size of 256 for all the experiments. The average was taken from 5 different runs, using the same seeds for different models.\\
\textbf{Auto-correlation and cross-correlation analysis:} For this analysis, a semi-supervised AV speech processing with the shallow MCC and baseline model is analysed. In this experimental setup, logFB audio features of dimension 22 and DCT visual features of dimension 50 were used \cite{adeel2019lip}. \\
\textbf{Resilience test:} Random processors were killed (set to zero) with a probability \textit{P}. To make the comparison as fair as possible, only processors from the convolutional layers were killed. Points were estimated from \textit{P=0 to 0.5} with steps of 0.025. We observed the whole testing dataset 50 times per point. The average/standard deviation was taken from the 5 runs of each model. It was observed that context-sensitive processors had significantly better resistance to processor damage. This is due to the fact that our model highlights the important features given the nature of the input, and does not look at the input processed features without any vis a vis importance or weight of the features.  Empirically, we observed that the quality of the reconstruction drastically decreased when going above the 0.01 error.\\
\textbf{Decoder details:} The decoder's initial layer is a fully connected layer followed by an (8,8,64) reshape. 
We apply four transpose convolutional transformations with 64, 32, 16 and 1 filters respectively. 
Kernel size is 3, the stride is 2 for all four convolutional steps.  
Activation function is ReLU. Batch normalization is added prior to each ReLU to enhance even further learning speed. The decoder's network is initialized with a glorot uniform distribution. 
This simple decoder is enough to achieve an almost perfect reconstruction when provided with the clean input in just a matter of a few updates (data not shown).
Thus, the final quality of the reconstruction is entirely dependent on the quality of the features provided by the encoder. 
Additional decoders required by the autoencoders have a similar structure.
\\\textbf{Acknowledgments}
This research was supported by the UK Engineering and Physical Sciences Research Council (EPSRC) Grant Ref. EP/T021063/1. We would like to acknowledge Professor Bill Phillips from the University of, Professor Peter König from the University of Osnabrück, Dr James Kay from the University of Glasgow, and  Professor Newton Howard from Oxford Computational Neuroscience for their help and support in several different ways, including reviewing our work, appreciation, and encouragement. \\
\textbf{Contributions}
AA conceived and developed the original idea, wrote the manuscript, and analysed the results. AA, MF, MR, and KA performed the simulations. \\
\textbf{Competing interests}
AA has a provisional patent application for the algorithm described in this article. The other authors declare no competing interests.
\\\textbf{Data availability} The data that support the findings of this study are available on request.
\bibliographystyle{IEEEtran}
\bibliography{NATURE.bib}

\onecolumn 
\pagebreak 
% \pagebreak 
% \pagebreak 
% \\
% \\

% \title{Supplementary Material}
\textbf{\Large Supplementary Material}
% \textbf{Supplementary Material}
\pagebreak 
\setcounter{figure}{0}
\renewcommand{\thefigure}{S\arabic{figure}}

\begin{figure*} [!]
	\centering
	\includegraphics[trim=0cm 0cm 0cm 0cm, clip=true, width=1\textwidth]{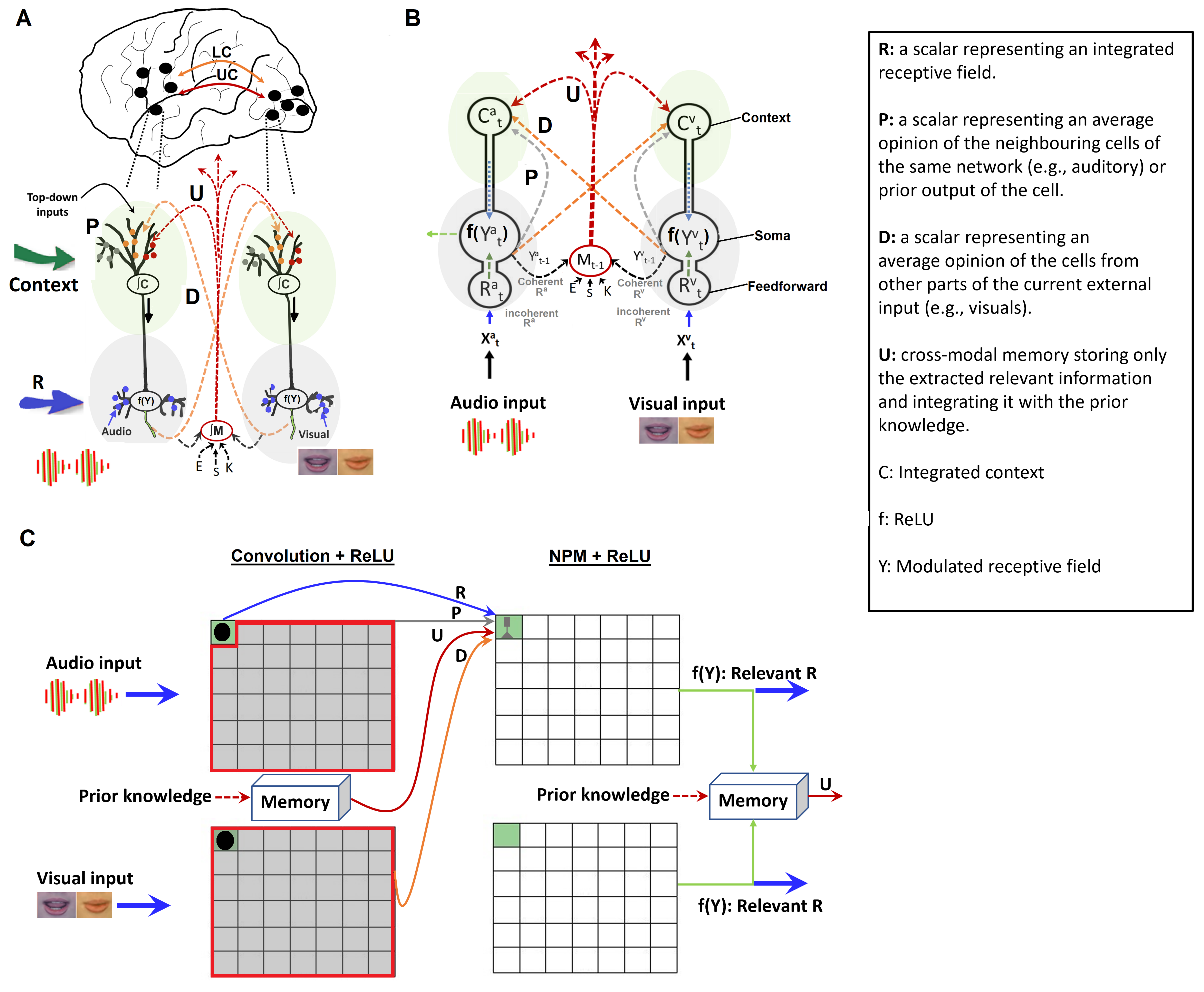}
	\caption{\textbf{Context-sensitive neural information processing: detailed information flow.} \textbf{(A)} Two-compartment two-unit circuit. The receptive field (R) in blue arrives at the basal. The local context (LC) (distal (D) in orange and proximal (P) in grey) and the universal context (U) in maroon arrive via synapses at the apical. U could explicitly be extended to the sources of inputs to include prior knowledge (K), emotions (E), and semantic knowledge (S). \textbf{(B)} Individual context-sensitive processors cooperate moment-by-moment via local and universal forms of context to separate coherent from conflicting signals via asynchronous modulatory transfer functions with the conditional probability of Y: $Pr(Y=1|R=r, C=c)=p(T(r,c))$, where \textit{p} is the half-Gaussian filter and T(r,c) is a continuous $\mathbb{R}^2$ function. The extracted coherent signals are recombined to extract synergistic memory signals. \textbf{(C)} Formation of contextual fields in a convolutional neural net. The convolutional block uses conventional point processors to generate R, P, D, and U, and the non-parametric modulation (NPM) block uses context-sensitive processors. Note that R in NMP block is non-parametric. } 
 \vspace{-1.2em}
\end{figure*}

% \begin{figure*} 
% 	\centering
% 	\includegraphics[trim=0cm 0cm 0cm 0cm, clip=true, width=1\textwidth]{tfsn1n.pdf}
% 	\caption{\textbf{Asynchronous modulatory transfer functions (AMTFs) (contour plots).} \textbf{(A-D)} Conventional AMTFs suggest that context modulates the input strength with which they transmit information about other inputs e.g., in these functions `R' is a driving force: when R is very weak, the cell's output is zero, and when it is strong, the output is high regardless of C. \textbf{(E)} In the proposed AMTF, C is a driving force. It overrules the typical dominance of R, therefore, the neural output is more dependent on the strength of C. In summary, the stronger the C, the stronger the relevance of R.}
%  \vspace{-1.2em}
% \end{figure*}

\begin{figure*} 
	\centering
	\includegraphics[trim=0cm 0cm 0cm 0cm, clip=true, width=1\textwidth]{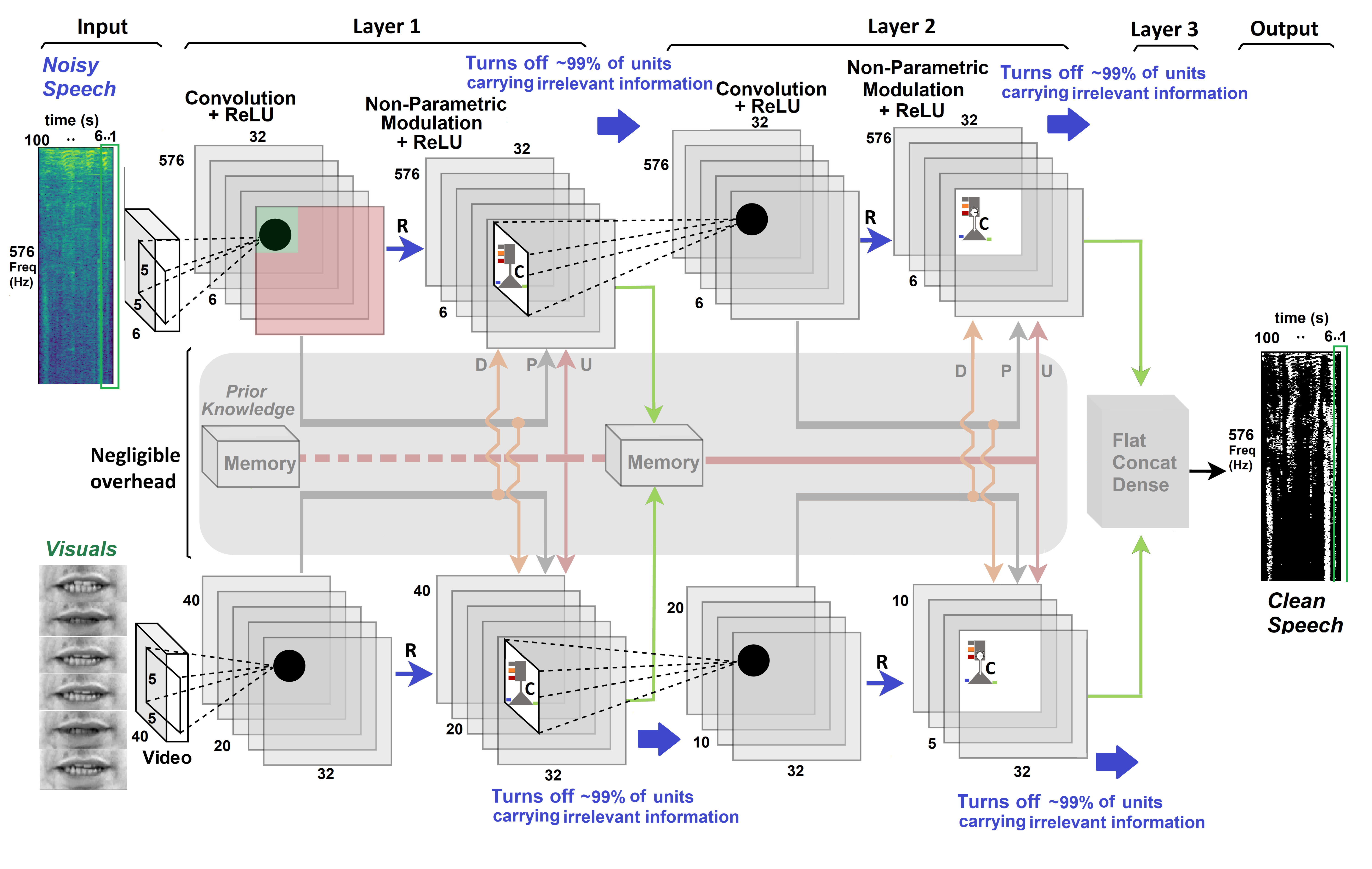}
	\caption{\textbf{Example context-sensitive deep information processing architecture:} the convolutional blocks use conventional point processors to generate R, P, D, and U. The non-parametric
modulation blocks, composed of context-sensitive processors, turn off 99\% of units carrying irrelevant information. }
 \vspace{-1.2em}
\end{figure*}

\begin{figure*} 
	\centering
	\includegraphics[trim=0cm 0cm 0cm 0cm, clip=true, width=1\textwidth]{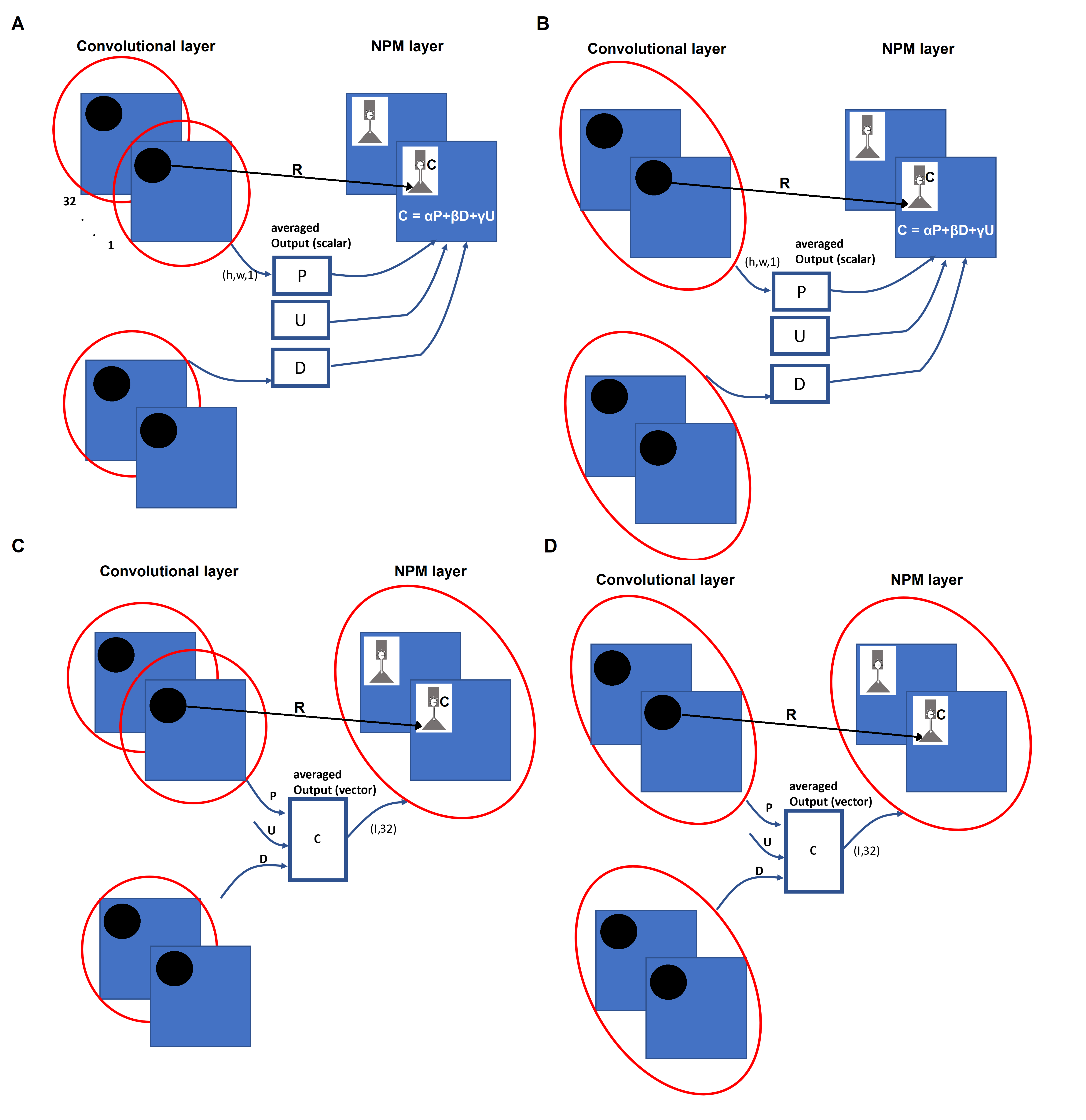}
	\caption{\textbf{A few possible configurations for context-sensitive deep information processing.} \textbf{(A)} Feature map-wise modulation: scalar contextual fields (CFs). In this case, each feature map (FM) is averaged and multiplied by a single  weight value (e.g., $\alpha$ for P, $\beta$ for D, and $\gamma$ for U) in the non-parametric modulation (NPM) layer. This configuration has an overhead of 32 parameter per CF.  \textbf{(B)} Feature map reduction: scalar CFs. In this case, 32 FMs are reduced, averaged, and multiplied by a single weight value.  This configuration has an overhead of 1 parameter per CF (P, D, and U) \textbf{(C)} Feature map-wise modulation: vector CFs. In this case, each FM is passed through an integrated contextual block (C) that comprises a mixture of convolutional and dense layers, and outputs 32 context values for 32 FMs (e.g., See Table 1). This configuration has an overhead of ~10.1K parameters per CF. \textbf{(D)} Feature map reduction: vector CFs. In this case, FMs are reduced and then passed through the C block that comprises a mixture of convolutional and dense layers, and outputs 32 context values for 32 FMs. This configuration has an overhead of ~10.1K parameters per CF.}
 \vspace{-1.2em}
\end{figure*}

\renewcommand{\thetable}{S\arabic{table}}
\begin{table*} 
	\centering
\includegraphics[trim=0cm 0cm 0cm 0cm, clip=true, width=1\textwidth]{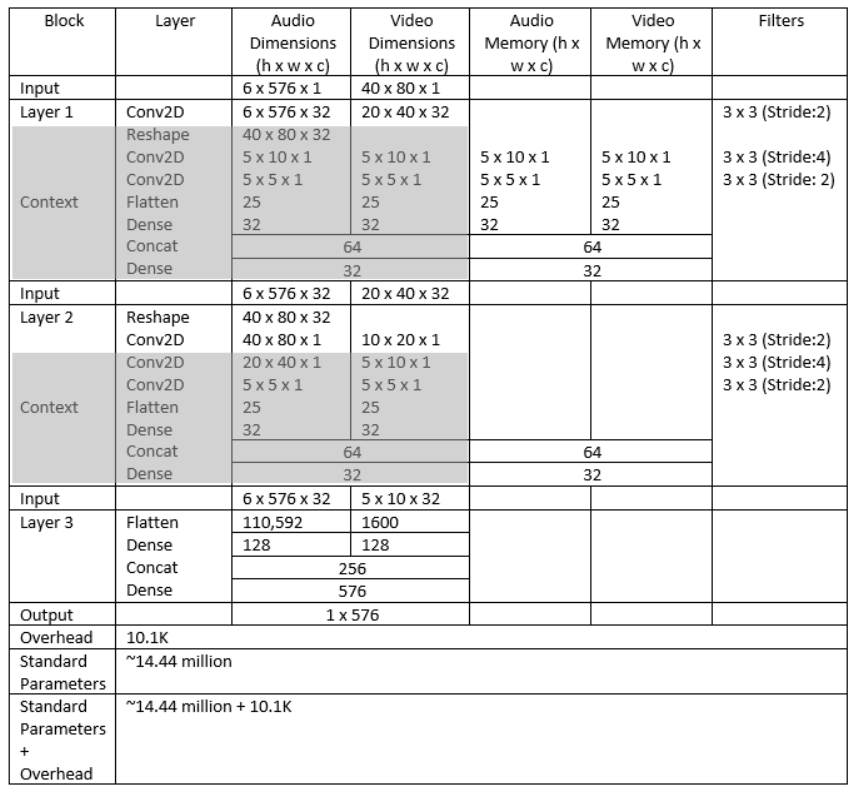}
	\caption{\textbf{Context-sensitive deep information processing architecture:} layer configuration and dimensions of the deep net used for analyses: feature map-wise point-wise modulation: vector contextual fields.}
 \vspace{-1.2em}
\end{table*}

\begin{table*} 
	\centering
\includegraphics[trim=0cm 0cm 0cm 0cm, clip=true, width=1\textwidth]{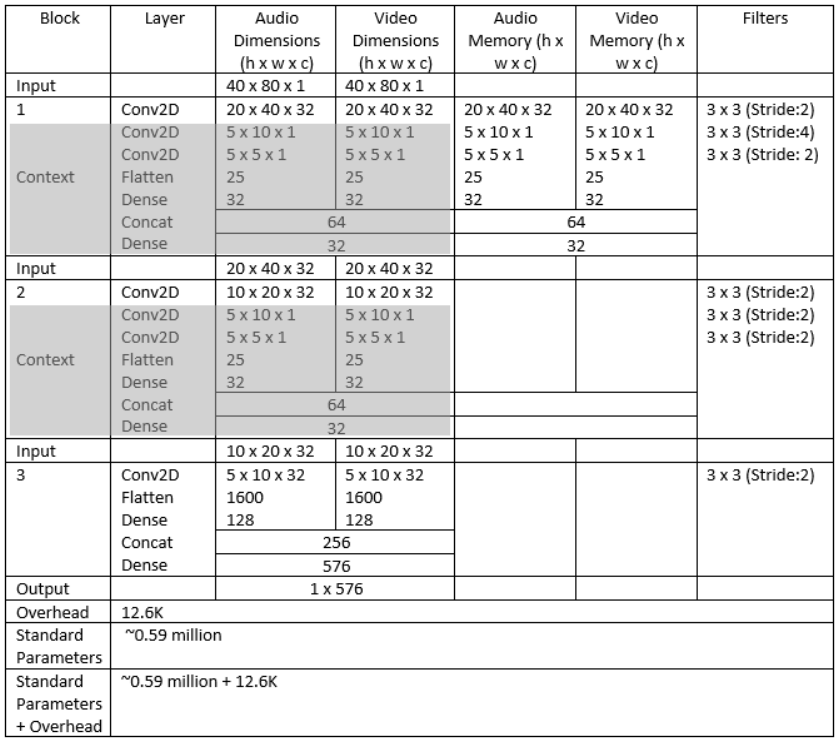}
	\caption{\textbf{Context-sensitive deep information processing architecture for reshaped speech signal (from 6$\times$576 to 40$\times$80)}.}
 \vspace{-1.2em}
\end{table*}

\begin{figure*} 
	\centering
	\includegraphics[trim=0cm 0cm 0cm 0cm, clip=true, width=0.5\textwidth]{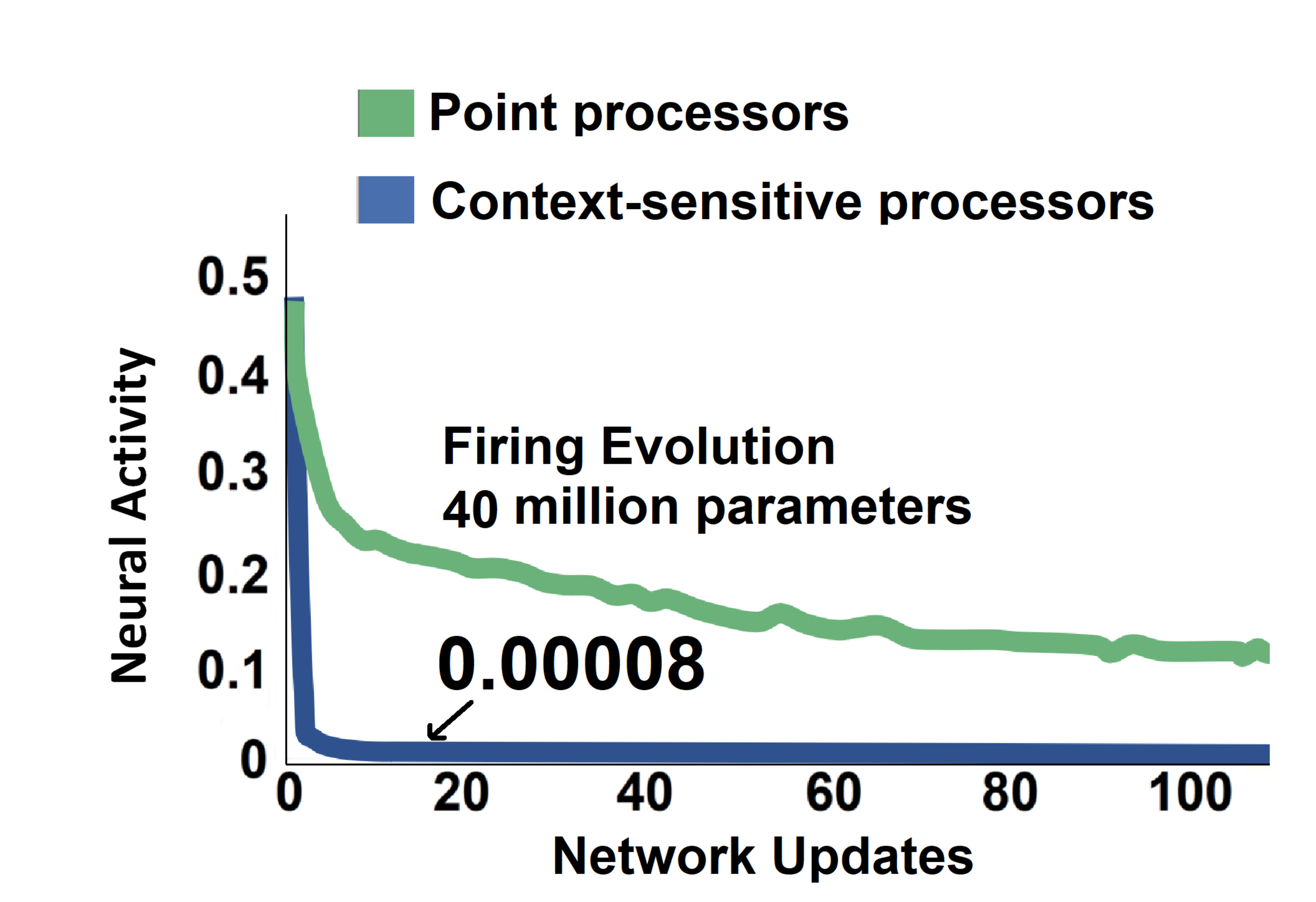}
	\caption{\textbf{Selective information processing: point processors vs context-sensitive processors} For
a larger deep model comprising 40 million parameters, the activity in context-sensitive processors reduces to ~0.008\% i.e., ~1250x less (per FF transmission) than the baseline. However, the reconstruction accuracy for context-sensitive processors and point processors drops to 85\% and 88\%,
respectively. In this case, more tuning and optimisation are required to search for Pareto-optimal.}
 \vspace{-1.2em}
\end{figure*}

\begin{figure*} 
	\centering
	\includegraphics[trim=0cm 0cm 0cm 0cm, clip=true, width=0.5\textwidth]{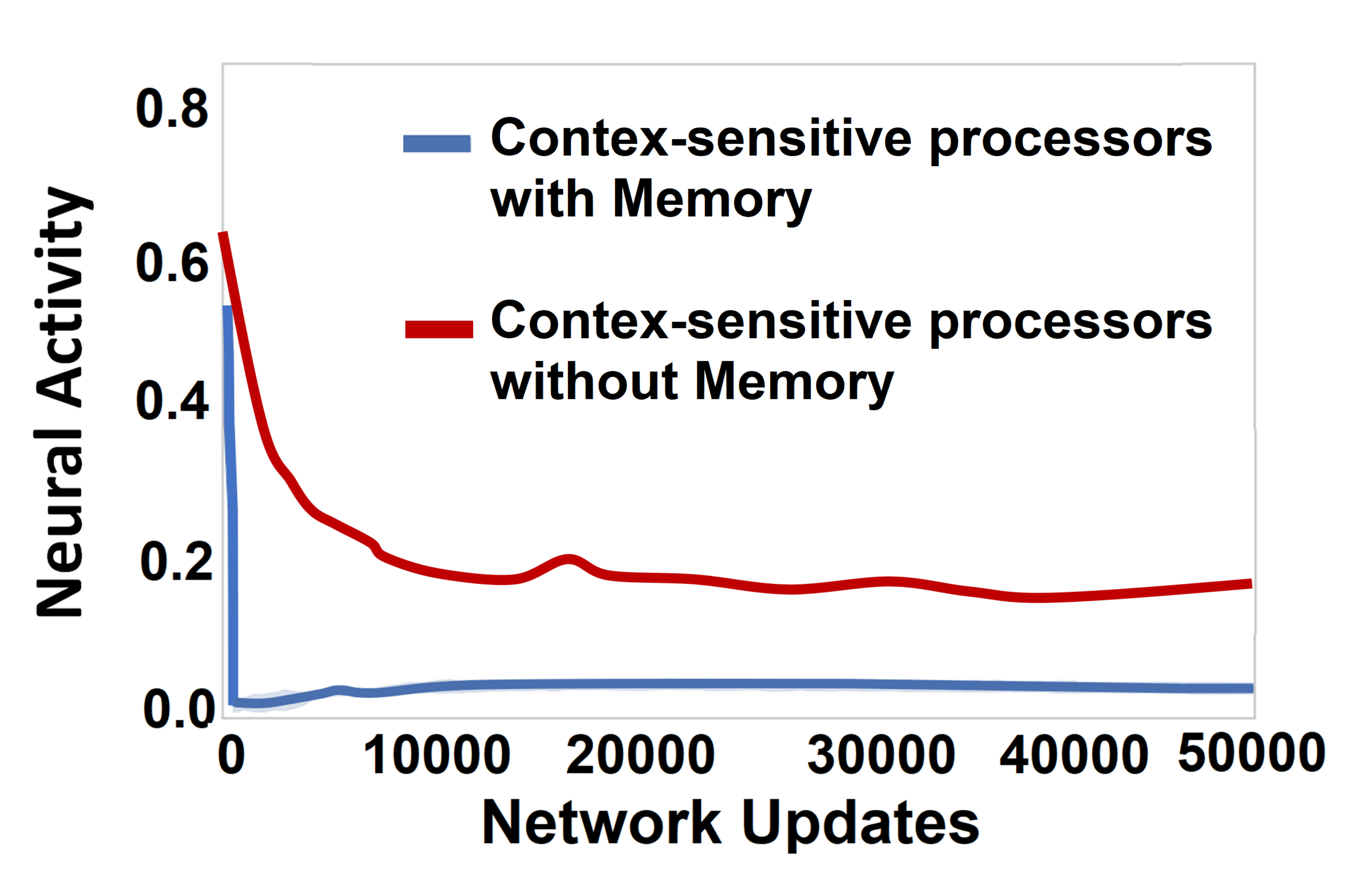}
	\caption{\textbf{Selective information processing: context-processors with memory vs. context-sensitive processors without memory.} For a  model comprising 14 million parameters, context-sensitive processors without memory reduce their activity but converge to a higher value. This behaviour shows that the higher the context, the higher the efficient information processing.}
 \vspace{-1.2em}
\end{figure*}

\begin{figure*}
    \centering
	\includegraphics[width=1\textwidth,keepaspectratio]{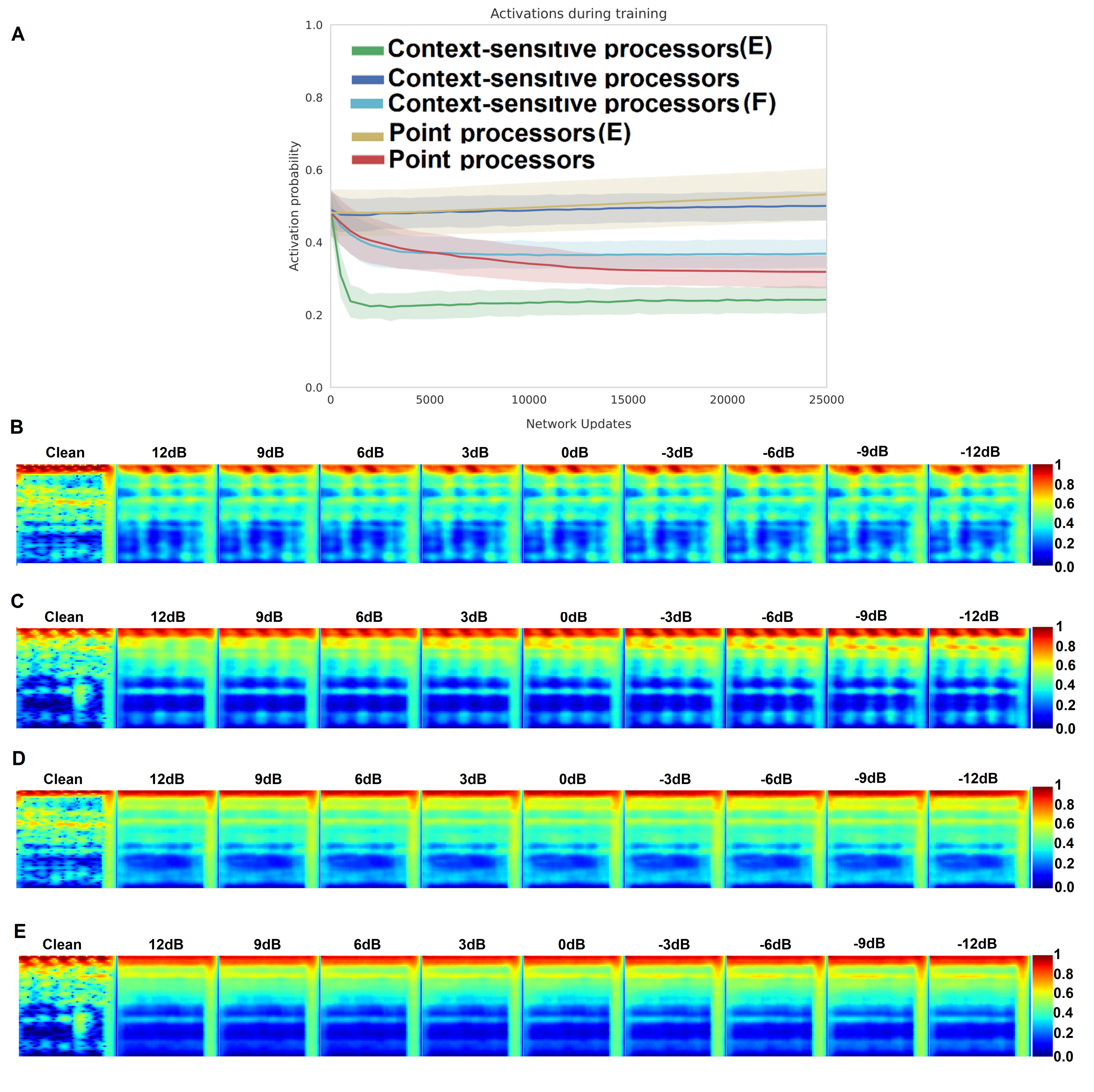}
	\caption{\textbf{Reconstructing high dimensional short-time Fourier transform.}   \textbf{(A)} Firing evolution: Context-sensitive processors quickly evolve to become highly sensitive to relevant information and become active only when the received information is important for the task at hand. Thus, the deep net, composed of context-sensitive processors, can separate clean speech from large amounts of noise using far fewer processors. Fast (F) represents deep network with higher learning rate. \textbf{(B-C)} Context-sensitive processors generalisation: clean-signal reconstruction for different levels of noise. \textbf{(D-E)} Point processors generalisation for different levels of noise: It is to be noted that context-sensitive processors capture high-frequency features more easily compared to the baseline.}
	\label{fig:deep_neural_activity}
 % \vspace{-1em}
\end{figure*}

\begin{figure*}
    \centering
	\includegraphics[width=1\textwidth,keepaspectratio]{ActivationsDistributionsn.pdf}
	\caption{\textbf{Neural activity convergence speed.} context-sensitive processors reach low neural activity 10X faster than the baseline model. For example, see row 2, column 3. The X-axis represents  processor's firing probability.}
	\label{fig:deep_neural_activity}
 % \vspace{-1em}
\end{figure*}

\begin{figure*} 
	\centering
	\includegraphics[trim=0cm 0cm 0cm 0cm, clip=true, width=0.5\textwidth]{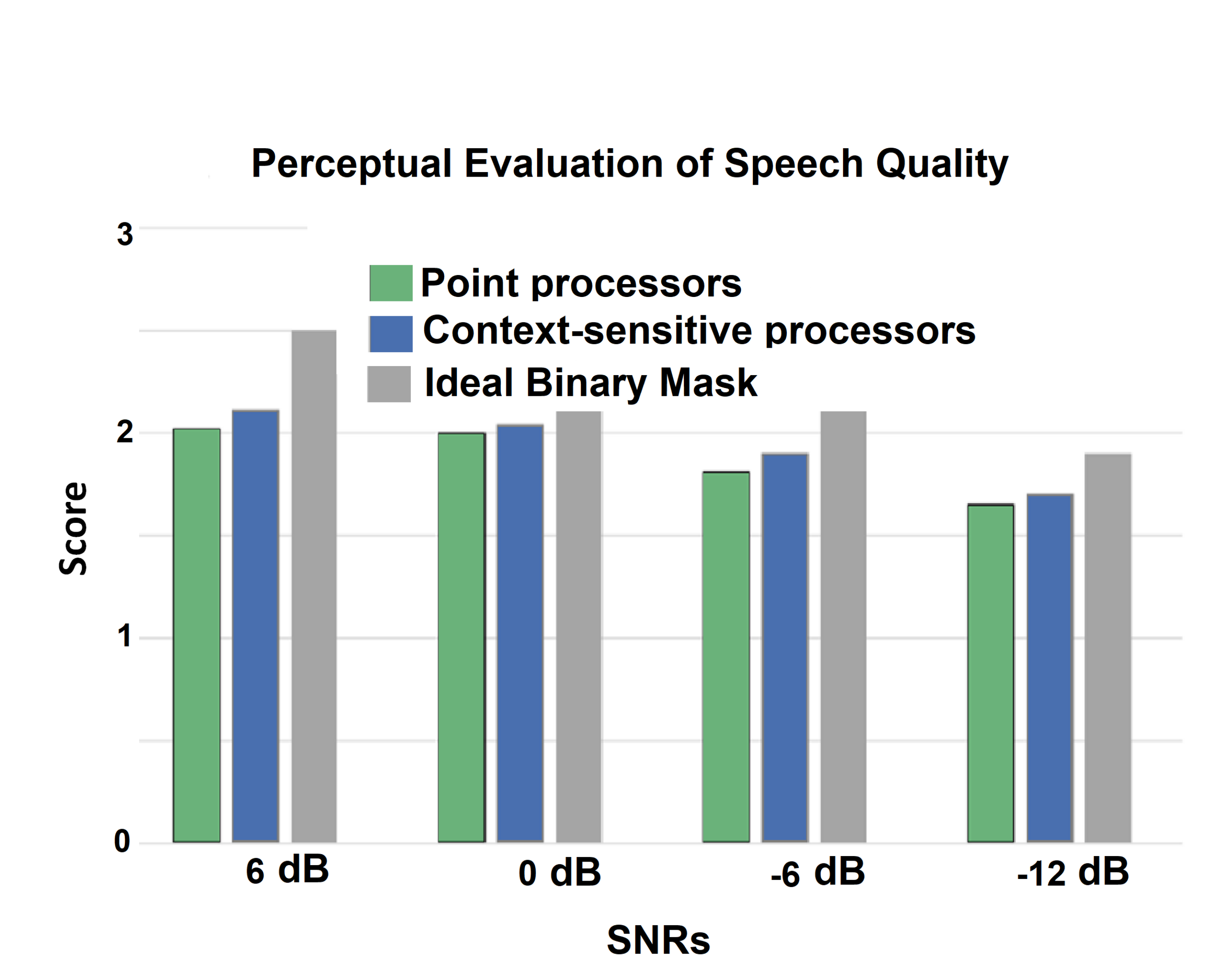}
\caption{\textbf{Perceptual evaluation of speech quality (PESQ)}. PESQ is objectively measuring the quality of re-synthesised speech for ideal binary mask estimation. }
 \vspace{-1.2em}
\end{figure*}

\begin{figure*} 
	\centering
	\includegraphics[trim=0cm 0cm 0cm 0cm, clip=true, width=\textwidth]{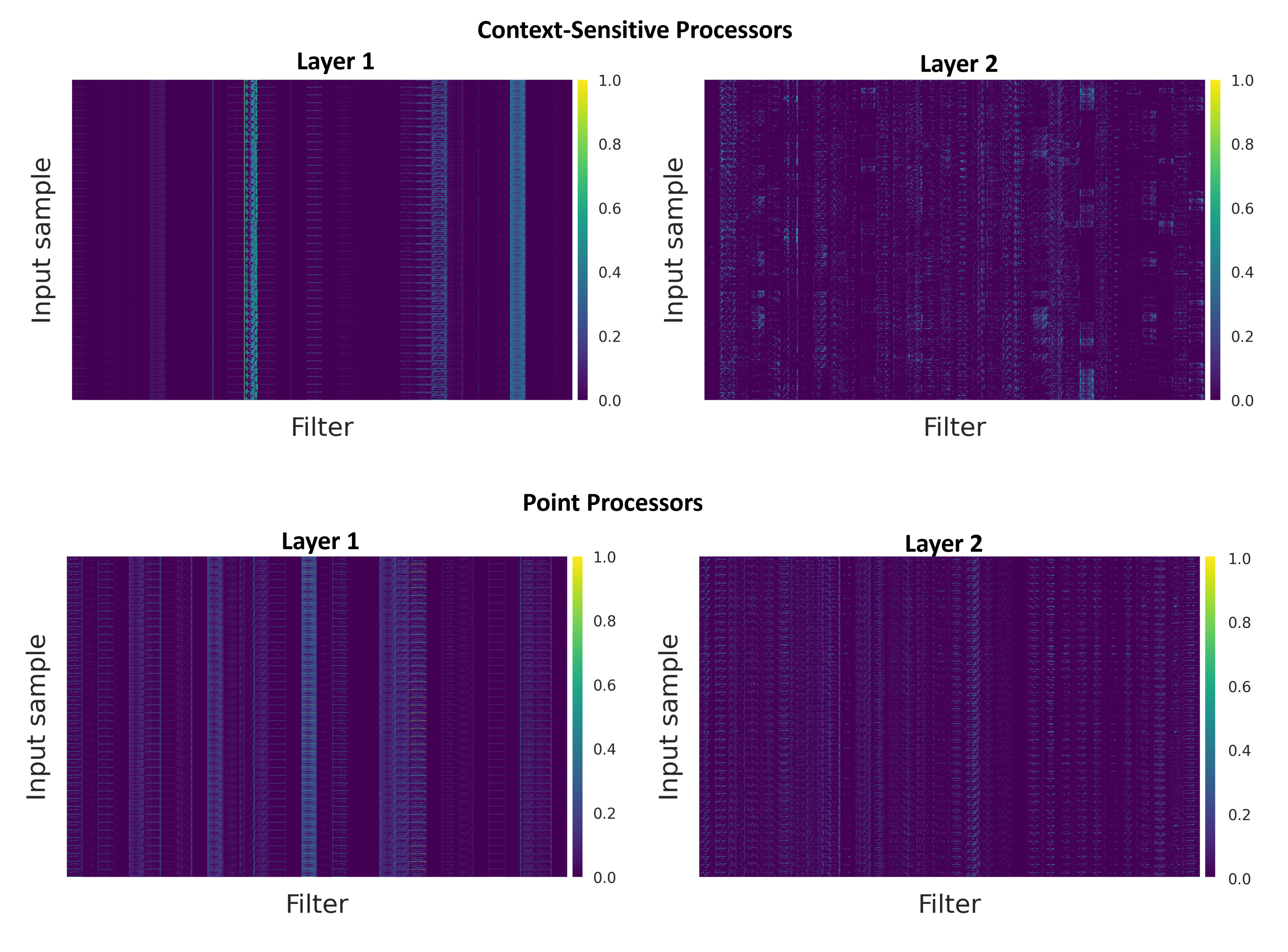}
	\caption{\textbf{Amplification and suppression of relevant and irrelevant FF signals, respectively for video blocks.} When processing visual information, context-sensitive processors, similar to audio processing, are restricting the transmission of irrelevant information to higher levels e.g.,  fewer  filters in Layer 1 and Layer 2 are active (indicating the most relevant information and significantly reducing the search space) as compared to the baseline.}
 	\label{fig:mi_mv_gaussian}
 \vspace{-1em}
\end{figure*}

\begin{figure*} 
	\centering
	\includegraphics[trim=0cm 0cm 0cm 0cm, clip=true, width=0.65\textwidth]{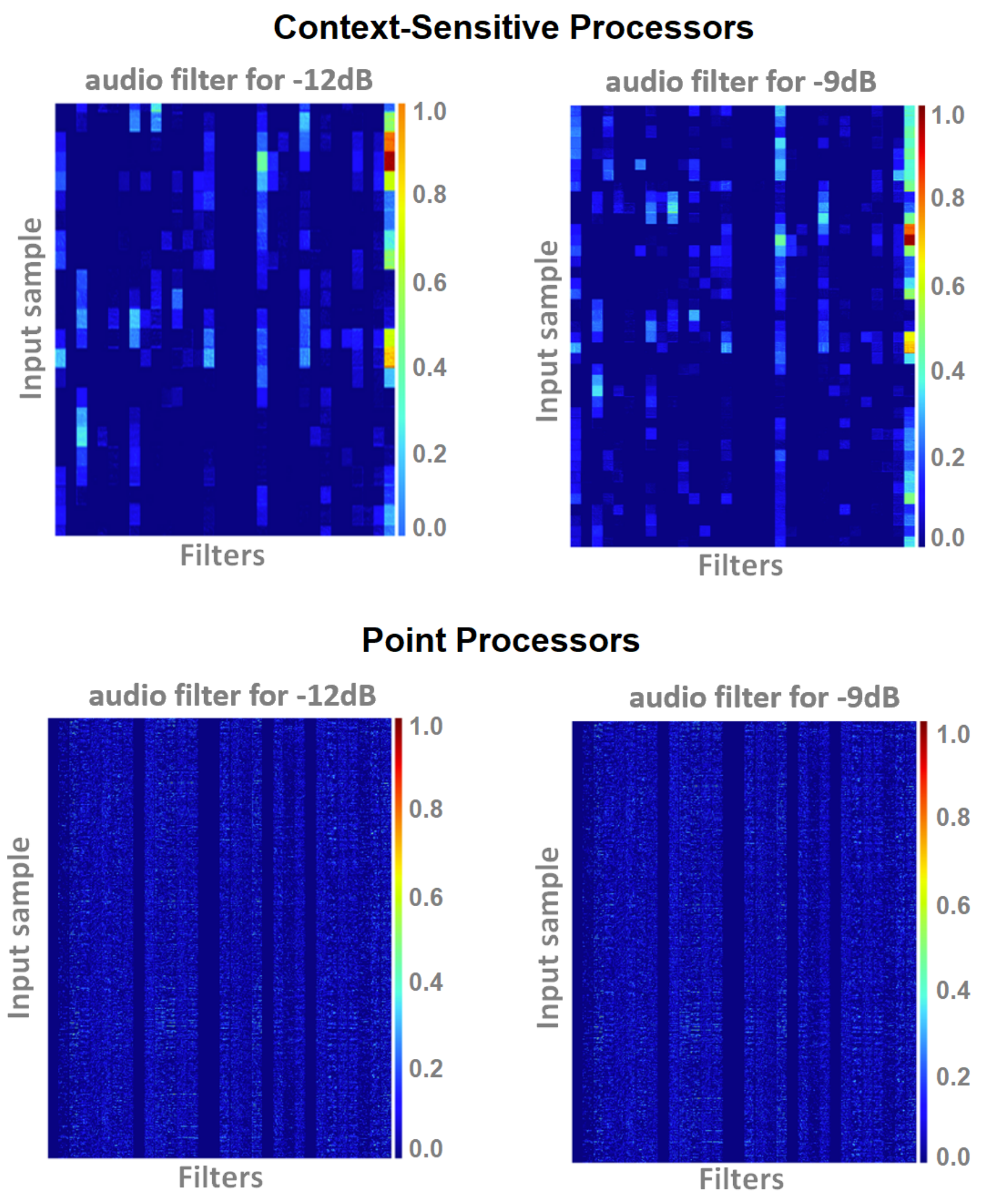}
	\caption{\textbf{Amplification and suppression of relevant and irrelevant FF signals, respectively for audio blocks for different SNRs.} It can be observed that different filters in MCC across the rows indicate what matters when. In contrast, the baseline treats each input equally, ignoring the variant information across the time. Note that context-sensitive processors use a full range of available frequency spectrum e.g., filters in red, green, blue, and orange to emphasise the level of relevance, whereas the irrelevant processors are off. }
 \vspace{-1.2em}
\end{figure*}

\begin{table}[htb!]
\centering \caption{Grid/ ChiME3 Corpus.} \label{}   \includegraphics[trim=0cm 0cm 0cm 0cm, clip=true, width=0.5\textwidth]{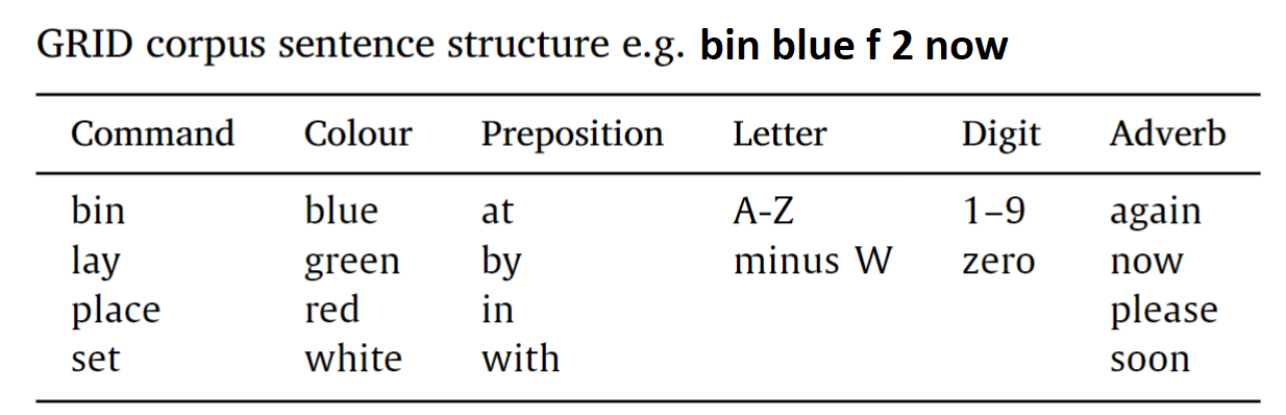}\end{table} 
% \begin{figure*}
%     \centering
% \begin{subfigure}{.75\textwidth}
%     \centering
%     \includegraphics[width=1\linewidth]{pre-pro1.pdf}  
%     \caption{Pre-processing}
%     \label{}
% \end{subfigure}
%         \begin{subfigure}{1\textwidth}
%         \centering
%         %\includegraphics[width=1\textwidth]{Figures/DeepMSE.pdf}
%         \includegraphics[width=1\textwidth]{pre-pro2.pdf}
%         \caption{Mask estimation based speech enhancement}
%          \end{subfigure}
%     \caption{\textbf{(a) Audio feature extraction:} the input audio signal is sampled at 16kHz and segmented into 75  frames each 84ms with 1350 samples per frame and stride of 13ms. Next, a hamming window and Fourier transformation is applied to produce the 622-bin power spectrum. Similarly, visual features are extracted from the Grid and ChiME3 corpora. Grid corpus videos are recorded at 25 fps whereas ChiME3 corpus is recorded at 75 fps. For visual features extraction, the video files are first processed to extract a sequence of individual frames. A cubic interpolation is used to match the visual frame rate with the audio frame rate of 75 frames per second. Afterwards, a FaceBlaze model is used as detector and Attention Mesh is used to identify  face landmarks and the lip-region. \textbf{(b) Mask estimation based speech enhancement:} prior audiovisual frames are used to incorporate temporal information. }
%     \label{}
%     \vspace{-1em}
% \end{figure*}

\begin{figure*} 
	\centering
	\includegraphics[trim=0cm 0cm 0cm 0cm, clip=true, width=\textwidth]{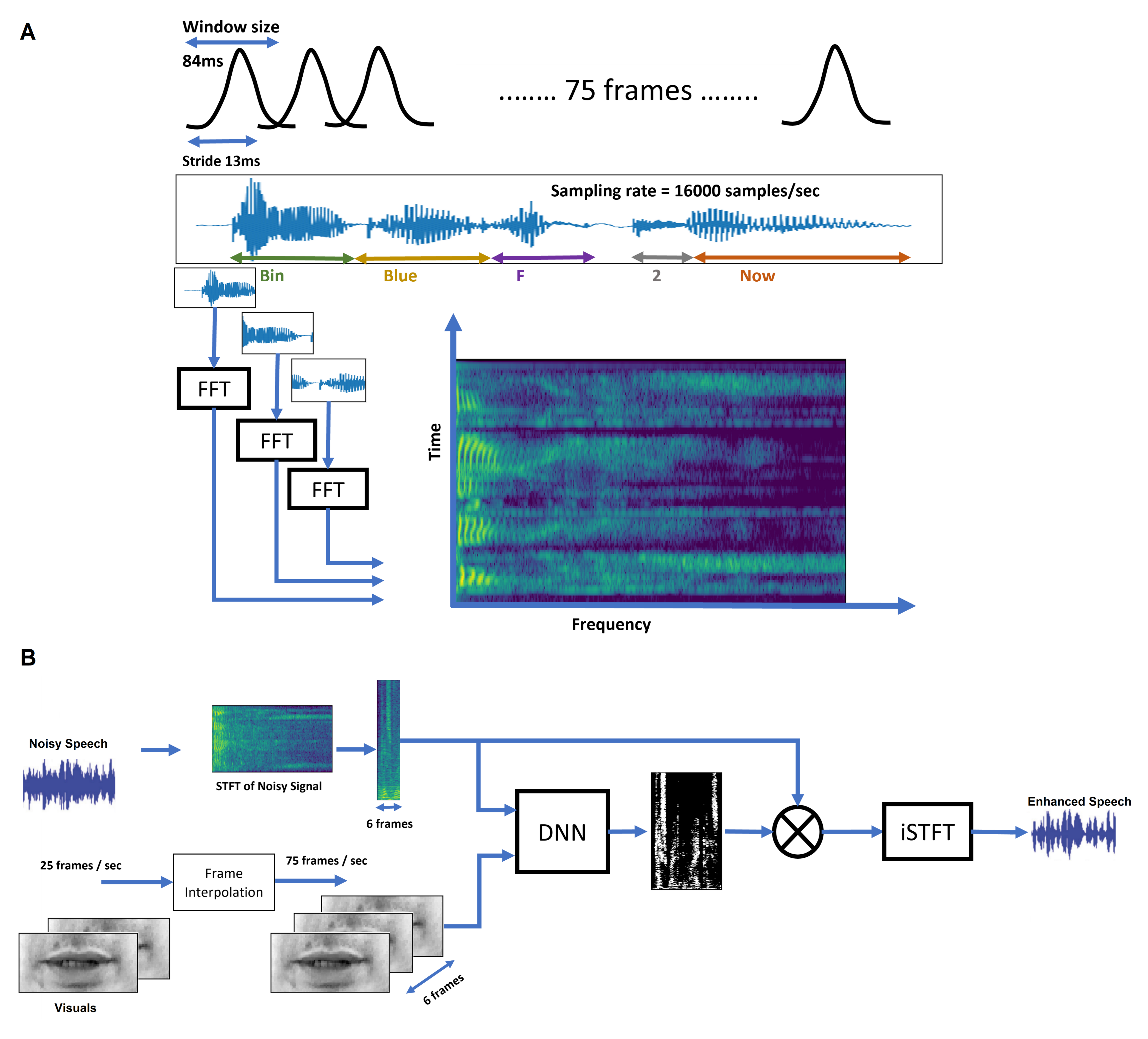}
	\caption{\textbf{Data pre-processing} \textbf{(A)} Audio feature extraction: the input audio signal is sampled at 16kHz and segmented into 75  frames each 84ms with 1350 samples per frame and stride of 13ms. Next, a hamming window and Fourier transformation is applied to produce the 622-bin power spectrum. Similarly, visual features are extracted from the Grid and ChiME3 corpora. Grid corpus videos are recorded at 25 fps whereas ChiME3 corpus is recorded at 75 fps. For visual features extraction, the video files are first processed to extract a sequence of individual frames. A cubic interpolation is used to match the visual frame rate with the audio frame rate of 75 frames per second. Afterwards, a FaceBlaze model is used as detector and Attention Mesh is used to identify  face landmarks and the lip-region. \textbf{(B)} Mask estimation based speech enhancement: prior audiovisual frames are used to incorporate temporal information.}
 	\label{fig:mi_mv_gaussian}
 \vspace{-1em}
\end{figure*}

\end{document}